\documentclass[10pt,twocolumn,letterpaper]{article}

\usepackage{cvpr}              % To produce the CAMERA-READY version
\usepackage[dvipsnames]{xcolor}

\definecolor{cvprblue}{rgb}{0.21,0.49,0.74}
\usepackage[pagebackref,breaklinks,colorlinks,citecolor=cvprblue]{hyperref}

\usepackage[utf8]{inputenc} % allow utf-8 input
\usepackage[T1]{fontenc}    % use 8-bit T1 fonts
\usepackage{hyperref}       % hyperlinks
\usepackage{url}            % simple URL typesetting
\usepackage{booktabs}       % professional-quality tables
\usepackage{amsfonts}       % blackboard math symbols
\usepackage{nicefrac}       % compact symbols for 1/2, etc.
\usepackage{microtype}      % microtypography
\usepackage{bm}
\usepackage{amsmath}

\usepackage{natbib}
\setcitestyle{numbers,square}
\usepackage{doi}
\usepackage{algorithm}
\usepackage{algpseudocode}
\definecolor{mygray}{gray}{0.95}

\usepackage{multirow}
\usepackage{colortbl}
\usepackage{makecell}

\usepackage{wrapfig}
\usepackage{parcolumns}

\usepackage{blindtext}
\usepackage[rightcaption]{sidecap}
\usepackage{caption}

\def\paperID{2017} % *** Enter the Paper ID here
\def\confName{CVPR}
\def\confYear{2024}

\title{DiffBIR: Towards Blind Image Restoration with Generative Diffusion Prior}

\author{Xinqi Lin$^{1,}$\thanks{} \quad Jingwen He$^{2,3,*}$ \quad Ziyan Chen$^{1,2}$  \quad Zhaoyang Lyu$^{2}$  \quad Bo Dai$^{2}$ 
 \, Fanghua Yu$^{1}$ \\ Wanli Ouyang$^{2,3}$  \quad Yu Qiao$^{2}$  \quad Chao Dong$^{1,2,}$\thanks{}\\
\textsuperscript{1}Shenzhen Institute of Advanced Technology, Chinese Academy of Sciences,\\ \textsuperscript{2}Shanghai AI Laboratory \\
\textsuperscript{3}The Chinese University of Hong Kong
\vspace{-1em}
}

\begin{document}
% \maketitle
\twocolumn[{%
    \renewcommand\twocolumn[1][]{#1}%
    \maketitle
    \thispagestyle{empty}
    \begin{center}
        \vspace{-1em}
        \centering
        \includegraphics[width=0.95\linewidth]{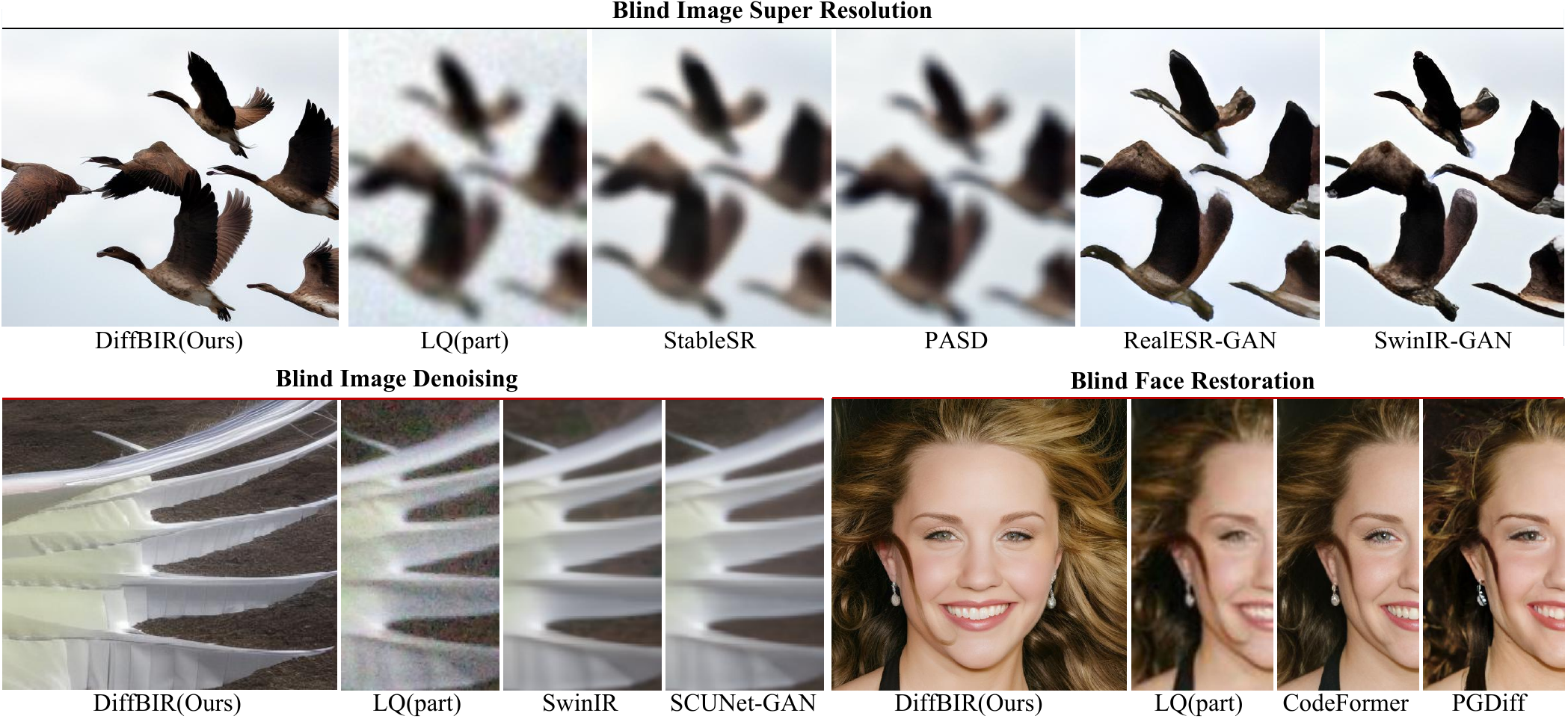}
        \vspace{-0.8em}
        \captionof{figure}{Comparisons of state-of-the-art methods and our DiffBIR for blind image super-resolution (BSR), blind image denoising (BID), and blind face restoration (BFR). (\textbf{Zoom in for best view})}
        \label{fig:teaser}
    \end{center}%
}]
{
  \renewcommand{\thefootnote}%
    {\fnsymbol{footnote}}
  \footnotetext[1]{Equal contribution}
  \footnotetext[2]{Corresponding author}
}
\begin{abstract}
\vspace{-1em}
We present DiffBIR, a general restoration pipeline that could handle different blind image restoration tasks in a unified framework. 
DiffBIR decouples blind image restoration problem into two stages:
1) degradation removal: removing image-independent content; 2) information regeneration: generating the lost image content. Each stage is developed independently but they work seamlessly in a cascaded manner.
In the first stage, we use restoration modules to remove degradations and obtain high-fidelity restored results. For the second stage, 
we propose IRControlNet that leverages the generative ability of latent diffusion models to generate realistic details. Specifically, IRControlNet is trained based on specially produced condition images without distracting noisy content for stable generation performance.
Moreover, we design a region-adaptive restoration guidance that can modify the denoising process during inference without model re-training, allowing users to balance realness and fidelity through a tunable guidance scale.
Extensive experiments have demonstrated DiffBIR's superiority over state-of-the-art approaches for blind image super-resolution, blind face restoration and blind image denoising tasks on both synthetic and real-world datasets. 
The code is available at \url{https://github.com/XPixelGroup/DiffBIR}.
\vspace{-2em}
\end{abstract}
\section{Introduction}
\label{sec:intro}

Image restoration aims at reconstructing a high-quality image from its low-quality observation. Typical image restoration problems, such as image denoising, deblurring and super-resolution, are usually defined under a constrained setting, where the degradation process is simple and known (\textit{e.g.}, bicubic downsampling). They have successfully promoted a vast number of excellent restoration algorithms \cite{srcnn, dncnn, swinir, IPT, uformer, restormer, hat}, but are born to have limited generalization ability. To deal with real-world degraded images, blind image restoration (BIR) comes into view and becomes a promising direction. The ultimate goal of BIR is to realize realistic image reconstruction on general images with general degradations. BIR does not only extend the boundary of classic image restoration tasks, but also has a wide practical application field (\textit{e.g.}, old photo/film restoration).

% \begin{figure}[t]
% \centering
% \subfloat[\small Visual comparison of blind image super-resolution (BSR) methods on real-world low-quality images.]{\includegraphics[width=1.0\textwidth]{figures/figure1_a.pdf}}\hfill
% \subfloat[Visual comparison of blind face restoration (BFR) methods on real-world low-quality face images.]{\includegraphics[width=1.0\textwidth]{figures/figure1_b.pdf}}\hfill

% \caption{\small Comparisons of DiffBIR and state-of-the-art BSR/BFR methods on real-world images.  Compared to BSR methods, DiffBIR is more effective to 1) generate natural textures; 2) reconstruct semantic regions; 3) not erase small details; 4) overcome severe cases. Compared to BFR methods, DiffBIR can 1) handle occlusion cases; 2) obtain satisfactory restoration beyond facial areas (\textit{e.g.}, headwear, earrings). (\textbf{Zoom in for best view})} \label{fig:teaser} \vspace{-1em}
% \end{figure}
% The research of BIR is still in its primary stage, thus requiring more explanations of its current state. 
% According to the problem settings, existing BIR problems can be roughly grouped into three research topics, namely 
% blind image super-resolution (BSR), zero-shot image restoration (ZIR) and blind face restoration (BFR). 

Typical BIR problems are blind image super-resolution (BSR), blind image denoising (BID), blind face restoration (BFR), etc.
% They all have achieved remarkable progress, but also have apparent limitations. 
BSR is initially proposed to solve real-world super-resolution problems, where the low-resolution image contains unknown degradations. 
The most popular solutions may be
BSRGAN \cite{bsrgan} and Real-ESRGAN \cite{realesrgan}. They formulate BSR as a supervised large-scale degradation overfitting problem. To simulate real-world degradations, a degradation shuffle strategy and high-order degradation modeling are proposed separately. Then the adversarial loss \cite{srgan, gans, esrgan, sngan, unetgan} and reconstruction loss are incorporated to learn the reconstruction process in an end-to-end manner.
They have demonstrated their great robustness in degradation removal for real-world super-resolution, but usually fail in generating realistic details due to the limited generative ability. 
% Furthermore, their degradation settings are limited to $\times4/\times8$ super-resolution, which is not complete for the BIR problem.
BID aims to achieve blind denoising \cite{scunet, cbdnet} for real-world noisy photographs, which usually contain various noises (\eg, dark current noise, short noise, and thermal
noise) due to the processing in real camera system. 
SCUNet~\cite{scunet} is the state-of-the-art method, which designs a practical noise degradation model to synthesize the noisy images, and adopts L1 loss as well as optional adversarial loss for training a deep denoiser model. Its solution is similar as BSR methods and thus has the same weakness.
BFR only focuses on blind restoration for face images. 
Due to a smaller image space, BFR methods (\eg, CodeFormer \cite{codeformer}, GFPGAN \cite{gfpgan}) could incorporate powerful generative facial priors (\eg, VQGAN \cite{vqgan}, StyleGAN \cite{stylegan2}) to generate faithful and high-quality facial details. They have achieved remarkable success in both academia and industry in recent years. Nevertheless, BFR assumes a fixed input size and restricted face image space, and thus cannot be applied to general images. 
% In conclusion, the above mentioned BIR tasks are faced with different challenges and have devised different solutions.

 % State-of-the-art methods can refer to CodeFormer \cite{codeformer} and VQFR \cite{vqfr}. They have a similar solution pipeline as BSR methods, but are different in the degradation model and generation network.

Recently, denoising diffusion probabilistic models (DDPMs \cite{ddpm}) have shown outstanding performance in image generation. DDRM \cite{ddrm}, DDNM \cite{ddnm}, and GDP \cite{gdp} incorporate the powerful diffusion model as the additional prior, thus having greater generative ability than GAN-based methods. With a proper degradation assumption, they can achieve impressive zero-shot restoration on classic IR tasks.  
However, the problem setting of zero-shot image restoration (ZIR) is not in accordance with BIR. Their methods can only deal with clearly defined degradations (linear or non-linear), but cannot generalize well to unknown degradations. 
In other words, they can achieve realistic reconstruction on general images, but not on general degradations. 
% Zero-shot image restoration is a newly emerged direction. Representative works are DDRM \cite{ddrm}, DDNM \cite{ddnm}, and GDP \cite{gdp}. They incorporate the powerful diffusion model as the additional prior, thus having greater generative ability than GAN-base methods. With a proper degradation assumption, they can achieve impressive zero-shot restoration on classic IR tasks. However, the problem setting of ZIR is not in accordance with BIR. Their methods can only deal with clearly defined degradations (linear or non-linear), but cannot generalize well to unknown degradations. In other words, they can achieve realistic reconstruction on general images, but not on general degradations. 

In this work, we aim to solve different BIR tasks in a unified framework. 
According to the review and analyses on recent progress in BIR tasks, we decouple the BIR problem into two stages: 1) \textit{degradation removal}: removing image-independent content; 2) \textit{information regeneration}: generating the lost image content. 
% For the first stage, we use different restoration modules to achieve degradation removal for different BIR tasks.
Considering that each BIR task corresponds to different degradation process and image dataset, we utilize different restoration modules to achieve degradation removal for each BIR task respectively. 
% Practically, the off-the-shelf MSE-based BIR models (\eg, BSRNet \cite{bsrgan} for BSR) are adopted.
% Although the degradation could be effectively removed, the restored images are still smooth and have incomplete image content, ranging from local semantic regions to high-frequency texture details. 
For the second stage, we utilize one generation module that leverages pre-trained text-to-image latent diffusion models \cite{sd} for generating faithful and visual-pleasing image content. By treating stage \uppercase\expandafter{\romannumeral2} as a conditional image generation problem, we have made some important observations that indicate
bad conditions, the original LQ images with distracting noises/artifacts, will disturb the generation process, causing unpleasant artifacts. Thus, we additionally train a MSE-based restoration module using simple degradation model with wide degradation ranges to produce reliable and diversified conditions. 
Furthermore, we propose IRControlNet to control the generative diffusion prior based on our produced conditions. Specifically, we use the pre-trained VAE encoder for condition encoding and follows ControlNet \cite{controlnet} to adopt an auxiliary and copied encoder for efficient add-on controlling. Our trained generation module remains effective and stable when combined with different restoration modules for different BIR tasks.
Moreover, a training-free controllable module is provided to trade-off between
\textit{fidelity} and \textit{quality}. 
Specifically, we introduce a training-free region-adaptive restoration guidance, which minimizes our designed region-adaptive MSE loss between the generated result and the high-fidelity guidance image at each sampling step through gradient-descent algorithm.
During guidance, the detected low-frequency regions are influenced more by the high-fidelity guidance image, while the high-frequency regions maintain more generative ability. 
Besides, a guidance scale can be tuned to achieve a smooth transition between two effects regarding \textit{fidelity} and \textit{quality}.

% An associated weight map is calculated based on the image gradients of the restored result in stage \uppercase\expandafter{\romannumeral1}. In this way, low-frequency regions are influenced more by high-fidelity
% guidance, while high-frequency regions maintain more generative ability. 
To sum up, the main contributions of this work are:
\begin{itemize}
    \item DiffBIR decouples BIR problem into two stages: restoration module for degradation removal, and generation module for lost information regeneration. With the two-stage design, DiffBIR is able to achieve the state-of-the-art performance for BSR, BFR, and BID tasks in a unified framework for the first time.
    \item We propose IRControlNet that leverages text-to-image diffusion prior for realistic image reconstruction. Comprehensive exploration on main components for generation module has been conducted, and IRControlNet proves to be a solid backbone for generation module in BIR tasks. 
    % \item We introduce a training-free region-adaptive restoration guidance, which modifies the denoising process at each timestep based on region-adaptive MSE loss between the generated result and the high-fidelity guidance image, providing a flexible trade-ff between \textit{quality} and \textit{fidelity}.
    \item We introduce a training-free controllable module -- region-adaptive restoration guidance that performs in sampling process, for achieving flexible trade-off between \textit{quality} and \textit{fidelity} for various user preferences.    
    % It minimizes our designed region-adaptive MSE loss between the generated result and the high-fidelity guidance image during sampling. A guidance scale can be tuned to  achieve flexible trade-off between \textit{quality} and \textit{fidelity}.
    %  through gradient-descent algorithm. 
\end{itemize}

\section{Related Work}
\label{sec:related_work}
% 1. BIR contains BSR (bsrgan, realesrgan)
% 2. BFR is a sub domain of BIR
\subsection{Blind Image Restoration}
\noindent\textbf{Blind Image Super-Resolution.} Latest advances \cite{bsr_survey} on BSR have explored more complex degradation models to approximate real-world degradations. In particular, BSRGAN~\cite{bsrgan} aims to synthesize more practical degradations based on a random shuffling strategy, and RealESRGAN~\cite{realesrgan} exploits "high-order" degradation modeling. SwinIR-GAN~\cite{swinir} uses the prevailing backbone Swin Transformer~\cite{swin_transformer} to achieve better image restoration performance. FeMaSR~\cite{femasr} formulates SR as a feature-matching problem based on pre-trained VQ-GAN~\cite{vqgan}.
% Although BSR methods can be useful to remove degradations in the real world, they are not good at generating realistic details. In addition, they typically assume the low-quality image input is downsampled by some certain scales (e.g. $\times 4/\times8$), which is limited for BIR problem.
Recently, the powerful Stable Diffusion has been leveraged for image restoration tasks. StableSR \cite{stablesr} designs a time-aware encoder to control the Stable Diffusion. PASD \cite{pasd} has proposed a PACA module, which could effectively inject the pixel-level condition information into diffusion prior and achieve higher fidelity. Although they have achieved great performance in real-world super-resolution, these methods require re-training for handling other image restoration tasks. 
% Latest advances in the BSR field, including BSRGAN~\cite{bsrgan} and RealESRGAN~\cite{realesrgan}, have explored to design of more complex degradation models to approximate real-world degradations. In particular, BSRGAN tries to synthesize more practical degradations based on a random shuffling strategy, and RealESRGAN exploits "high-order" degradation modeling to stimulate the real degradation generation process. Although BSR methods can be useful to remove degradations in the real world, they typically assume the low-quality image input is downsampled from a certain scale. As a task covering a larger scope compared to BSR, BIR restores an input with real-world degradation with an arbitrary upsampling scale. Specifically, blind face restoration (BFR), a sub-domain of BIR, has drawn increasing attention. With the development of Vision Transformer (ViT) \cite{vit1, vit2}, many works successfully apply this popular backbone to extract representative features from face images \cite{restoreformer, faceformer, codeformer}. Meanwhile, recent methods \cite{codeformer, vqfr} based on Vector-Quantized dictionary learning \cite{vqvae, vqgan} demonstrated their superiority in restoring degraded faces in the wild.

\noindent\textbf{Blind Face Restoration.}
% 1. gan prior (PULSE, DGP-Exploiting deep generative prior for versatile image restoration and manipulation), ddpm prior (ddrm, ddnm, gdp)
As a specific sub-domain of general images, the face image typically carries more structural information. 
% Early attempts utilize geometric priors (e.g. facial parsing maps~\cite{face_geo_prior_1}, facial landmarks\cite{fsrnet, face_geo_prior_2}, and facial component heatmaps~\cite{yu2018face}) or reference priors\cite{XiaomingLi2018LearningWG, XiaomingLi2020EnhancedBF, XiaomingLi2020BlindFR, BerkDogan2019ExemplarGF} as auxiliary information to guide the face restoration process. 
Recent BFR approaches mainly incorporate powerful generative priors to reconstruct faces with great realness. Representative GAN-prior-based methods~\cite{gfpgan, gpen, gcfsr, glean} have demonstrated their capability in achieving both high-quality and high-fidelity face reconstruction. State-of-the-art works~\cite{codeformer, vqfr, restoreformer} introduce the HQ codebook to generate surprisingly realistic face details by exploiting Vector-Quantized (VQ) dictionary learning~\cite{vqvae, vqgan}. Latest advances~\cite{difface, dr2, pgdiff} leverage the powerful generative capability of diffusion prior and achieve high-quality and robust face restoration. However, all these methods can only achieve good performance for face images, while the general images are beyond their scopes.
% TODO: what is ~\cite{} ?

\noindent\textbf{Blind Image Denoising.}
Blind image denoising (BID) aims to handle unknown noise levels/types. Several attempts have been made to solve the BID problem. Among them, DnCNN~\cite{dncnn}, as an end-to-end deep CNN, is proposed to handle Gaussian denoising with multiple noise levels. 
GCBD~\cite{gcbd} leverages generative adversarial networks (GAN) for noise modeling. CBDNet~\cite{cbdnet} uses a more realistic noise model to synthesize low-quality data and incorporates real-world noisy-clean image pairs.  VDNet~\cite{vdnet} proposes to implement noise estimation and denoising simultaneously based on the variational denoising network. Although the above methods have shown great ability in removing unknown noises, they usually produce smooth results.

% \vspace{-1em}
\subsection{Zero-shot Image Restoration.}
ZIR aims to achieve image restoration by leveraging a pre-trained prior network in an unsupervised manner. Earlier works \cite{zero_shot_gan_based1, zero_shot_gan_based2, pulse, dgp} mainly concentrate on searching a latent code within a pre-trained GAN's latent space. Recent advancements in this field embrace the utilization of DDPMs \cite{ddpm, songyang1, songyang2, sd, dalle2, imagen}. DDRM \cite{ddrm} introduces an SVD-based approach to handle linear image restoration tasks efficiently. Meanwhile, DDNM \cite{ddnm} analyzes the range-null space decomposition of a vector theoretically and then designs a sampling schedule based on the null space. Inspired by classifier guidance \cite{beatsgan}, GDP \cite{gdp} introduces a more convenient and effective guidance approach, in which the degradation model can be estimated during inference. Although these works contribute to the advancement of zero-shot image restoration techniques, ZIR methods still cannot achieve satisfactory restoration results in low-quality images from the real world.

% Image prior from a pre-trained generative model provides extra information derived from real-world image distribution, which therefore serves well for image restoration with inputs missing high-quality real-world details. Early attempts encapsulate the image prior in the generative adversarial network (GAN). The GAN-prior-based methods have demonstrated their applicability for restoration tasks in both face image \cite{pulse, gfpgan, gpen, gcfsr, glean} and general image \cite{bsrgan, realesrgan, dgp} scenarios. As an emerging generative model, denoising diffusion probabilistic models (DDPMs) have exhibited impressive generative capabilities and diversity on top of GAN. Inspired by the nature of the diffusion process, many recent works leverage pre-trained DDPM as an effective prior for image restoration using degraded low-quality inputs as guidance \cite{sr3, diff-face}, including zero-shot image restoration approaches \cite{ddrm, ddnm, gdp}.

\section{Method}
\subsection{Motivation and Framework}
\label{sec:motivation}
In this work, we aim to exploit a powerful generative prior to solve BIR problem. Generative diffusion prior has demonstrated its effectiveness in conditional image generation \cite{controlnet} through enabling condition inputs, such as edge and segmentation maps.
This provides a potential solution for BIR problem, that is to regard it as conditional image generation and directly utilize the LQ images as condition inputs. However, low-quality image domain is vast and complex, thus the corresponding condition information is extremely diversified. More importantly, as the degradation and content information of LQ images are entangled, directly treating them as control signals will cause instability and induce artifacts.
\begin{figure}
    % \vspace{-1.5em}
    \includegraphics[width=\linewidth]{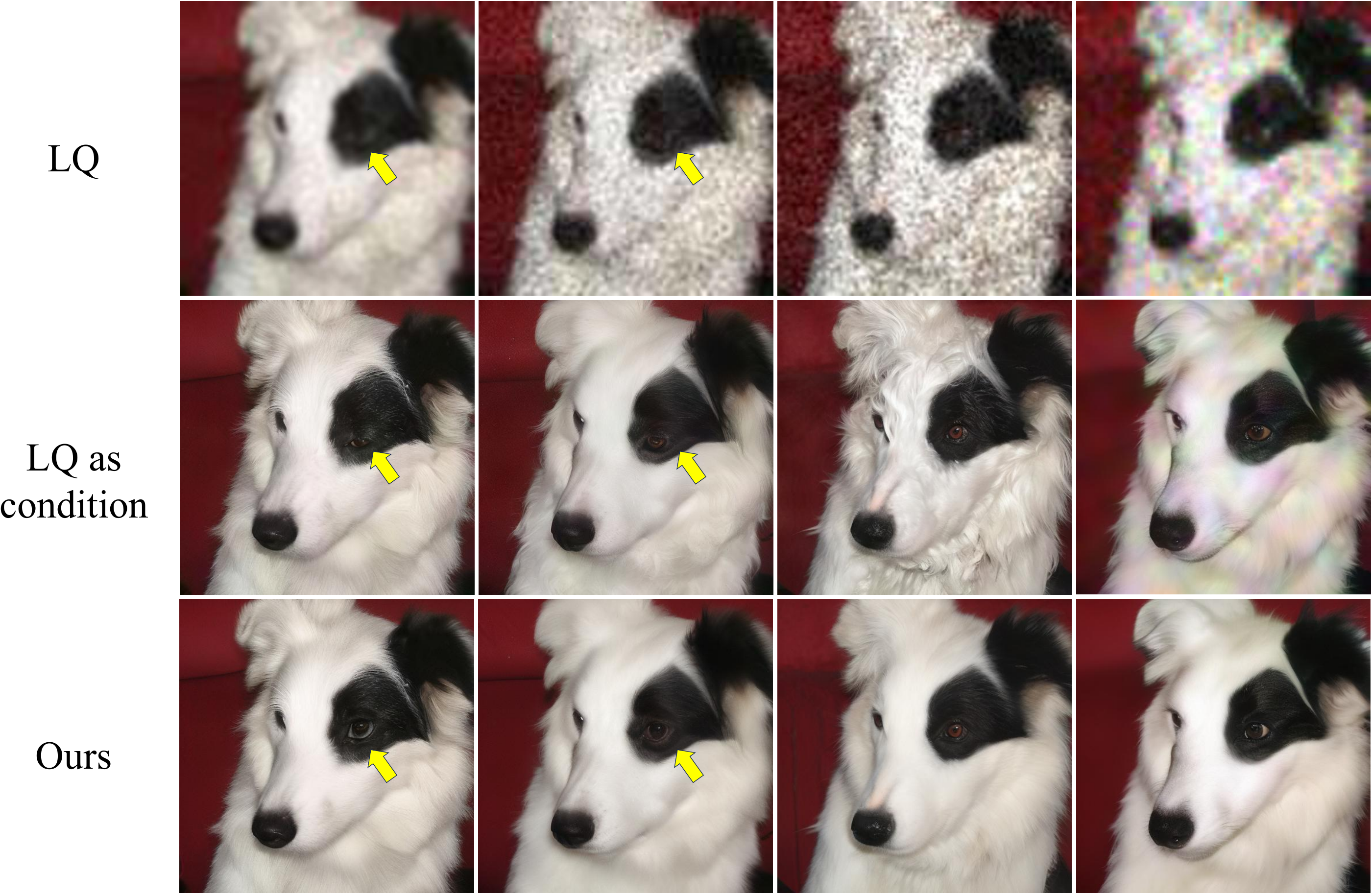}
    \vspace{-1em}
    \caption{The effects of condition information on generated results. The 2nd row shows that directly using LQ images as conditions causes unpleasant artifacts induced by different degradations (Gaussian, speckle, Poisson, and JPEG compression noises). While our DiffBIR's two-stage pipeline is more stable (see 3rd-row). }
    \label{fig:observation_RM}
    \vspace{-1.5em}
\end{figure}

% \begin{figure*}[h!]
\begin{figure*}[htbp]
    % \vspace{-0.5em}
    \centering
    \includegraphics[width=\linewidth]{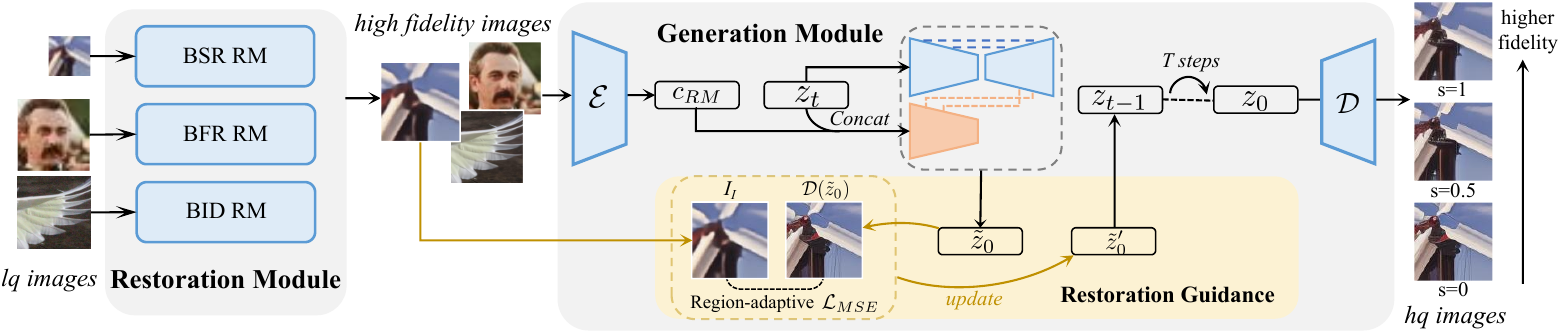}
    \vspace{-1.5em}
    \caption{The two-stage pipeline of DiffBIR. 1) Restoration Module (RM) for degradation removal; 2) Generation Module (GM) for realistic image reconstruction with optional region-adaptive restoration guidance for a trade-off between \textit{quality} and \textit{fidelity}.}
    \label{fig:pipeline}
    \vspace{-1em}
\end{figure*}

As presented in Fig. \ref{fig:observation_RM}, the LQ images are degraded with different types of noises based on the same HQ image. We train a generation module with synthesized LQ images as the conditions, and obtain the corresponding results. It is observed that the degradation indeed has an effect on the produced results: different unpleasant artifacts are generated due to the degradation difference. Since the training is not explicitly guided to distinguish the content information from the degraded image, the generation process is disturbed by the unreliable condition information.

% \begin{figure}[htbp]
%     \centering
%     \includegraphics[width=1\linewidth]{figures/observation_RM.pdf}
%     \vspace{-1em}
%     \caption{\small The effects of condition information on generated results. 
%     The 2nd row shows that directly using LQ images as conditions will cause unpleasant artifacts induced by different degradations (Gaussian, speckle, Poisson, and JPEG compression noises). While our DiffBIR's two-stage pipeline is more stable (see 3rd-row).}
%     \label{fig:observation_RM}
%     \vspace{-0.5em}
% \end{figure}

According to our observations and analyses, 
we adopt a general two-stage pipeline for BIR tasks, which contains restoration modules for removing image-independent degradation, and one generation module that only focuses on image content regeneration.
These two stages are decoupled and optimized independently. In this way, we can use any off-the-shelf/self-trained restoration module to address the challenging degradation removal for different BIR tasks. More importantly, the generation is only conditioned on the image content of LQ input, thus it will not be disturbed by degradation. This two-stage pipeline provides a flexible, stable, and unified solution to BIR problem.
% The first stage could be achieved by IR backbones that
% is trained with MSE loss and synthetic data. As for the generation module, we train an additional restoration module to produce reliable and 
% Specifically, the restoration modules adopt low-level backbones, and are trained with specially designed degradation models for different BIR tasks using MSE loss. While the generation module leverages the generative diffusion prior based on 
Besides, a training-free controllable module is introduced to achieve fidelity-quality trade-off by region-adaptive restoration guidance in the sampling process. The whole pipeline is illustrated in Fig. \ref{fig:pipeline}.

\subsection{Restoration Module}
\label{sec:method1}

% \begin{wrapfigure}{r}{0.5\textwidth}
%     % \vspace{-1.5em}
%     % \centering
%     \includegraphics[width=0.95\linewidth]{eccv_figures/RM.pdf}
%     % \vspace{-0.5em}
%     % \caption{\small This is RM.}
%     \label{fig:arch}
%     % \vspace{-1.5em}
% \end{wrapfigure}
% \begin{minipage}{0.48\linewidth}
%   \captionof{figure}{This is a caption for the image on the left.}
% \end{minipage}

In the first stage, we aim to remove distracting degradations of low-quality images without generating any new content for different BIR tasks. 
Note that each BIR task has its own characteristics in terms of degradation process and image dataset. For instance, BID methods should especially consider processed camera sensor noises, while BFR methods only focus on restoring low-quality face images. Therefore, we use separate restoration modules instead of a general one for different BIR tasks to maintain their expertise. 
% The recent state-of-the-art methods for BSR, BFR, and BID tasks all handle degradation removal based on specially designed degradation models. Their trained models with synthesized image pairs have obtained robust performance in real-world settings. 
In this work, we directly adopt the off-the-shelf BIR models trained with MSE loss as the restoration modules. 

% As mentioned in Section \ref{sec:motivation}, training a stable generation module requires reliable conditions. Besides, the generation module should be able to process the restored results from stage \uppercase\expandafter{\romannumeral1}.
As mentioned in Section \ref{sec:motivation}, training a stable generation module requires reliable conditions.
% To this end, we additionally train a restoration module (RM) just to produce appropriate condition images.
To this end, we additionally train a restoration module (RM) to produce appropriate condition images for training generation module.
Specifically, this RM is trained with classic degradation model and MSE loss:
\begin{equation}
    I_{RM} = \texttt{RM}(I_{lq}), \\
    \,\,\,\mathcal{L}_{RM} = ||I_{RM}-I_{hq}||^2_2,
\end{equation}
where $I_{hq}$, $I_{lq}$, and $I_{RM}$ denote the high-quality image, the synthesized low-quality counterpart, and the restored image, respectively.
Note that the degradation range is set to large since we desire to generate sufficiently diversified condition images. This will improve the overall generative capacity of the generation module (see Section \ref{sec:ablation3}). Please refer to Appendix for implementation details.
This naively trained RM performs as a condition preprocessing for the generation module, and it will be discarded during inference as it cannot handle complex degradations in real-world scenarios.

\subsection{Generation Module}
\label{sec:method2}
\noindent\textbf{Preliminary: Stable Diffusion.}
We implement our method based on the large-scale text-to-image latent diffusion model, Stable Diffusion.
To achieve better efficiency and stabilized training, Stable Diffusion pretrains an autoencoder \cite{kingma2013auto} that converts an image $x$ into a latent $z$ with encoder $\mathcal{E}$ and reconstructs it with decoder $\mathcal{D}$. 
Both diffusion and denoising processes are performed in the latent space.
In diffusion process, Gaussian noise with variance $\beta_t\in(0, 1)$ at time $t$ is added to the encoded latent $z = \mathcal{E}(x)$ to produce the noisy latent:
\begin{equation}
    z_t=\sqrt{\bar{\alpha}_t}z+\sqrt{1-\bar{\alpha}_t}\epsilon,
\end{equation}
where $\epsilon\sim\mathcal{N}(0, \textbf{I})$, 
$\alpha_t = 1 - \beta_t$ and $\bar{\alpha}_t = \prod^{t}_{s=1}\alpha_s$. When $t$ is large enough, the latent $z_t$ is nearly a
standard Gaussian distribution.
A network $\epsilon_{\theta}$ is learned by predicting the noise $\epsilon$ conditioned on $c$ (\textit{i.e.}, text prompts) at a randomly picked time-step $t$. The optimization of the latent diffusion model is defined as follows:
% \begin{equation}
%     \mathcal{L}_{ldm} = \mathbb{E}_{z, c, t, \epsilon}[||\epsilon-\epsilon_{\theta}(z_t=\sqrt{\bar{\alpha}_t}z+\sqrt{1-\bar{\alpha}_t}\epsilon, c, t)||^2_2],
% \end{equation}
\begin{equation}
    \mathcal{L}_{ldm} = \mathbb{E}_{z, c, t, \epsilon}[||\epsilon-\epsilon_{\theta}(\sqrt{\bar{\alpha}_t}z+\sqrt{1-\bar{\alpha}_t}\epsilon, c, t)||^2_2],
\end{equation}
where $x,c$ are sampled from the dataset and $z=\mathcal{E}(x)$, $t$ is uniformly sampled and $\epsilon$ is sampled from the standard Gaussian distribution.

\begin{figure*}
    % \vspace{-2em}
    \centering
    \includegraphics[width=\linewidth]{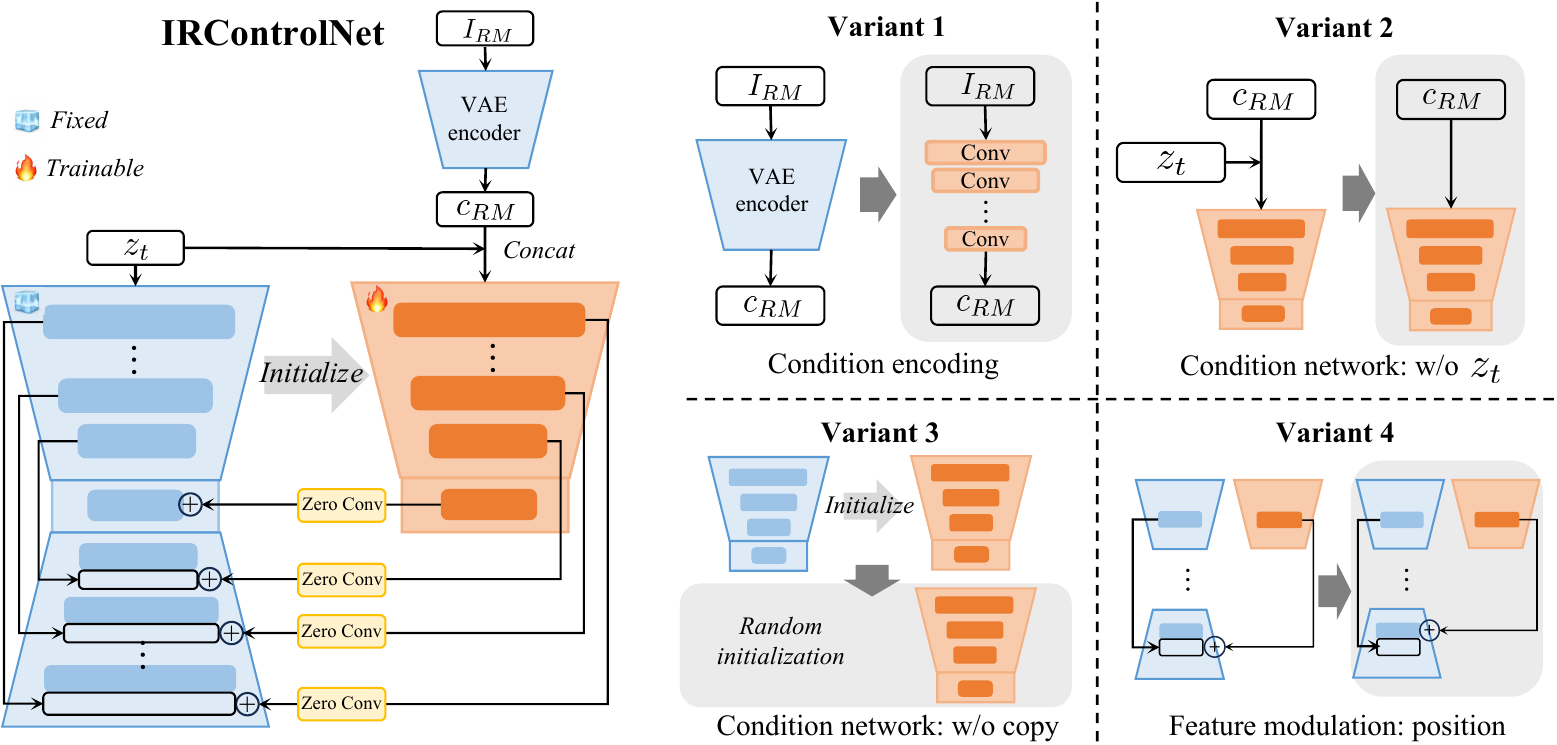}
    \vspace{-1.5em}
    \caption{Architectures of our IRControlNet and four model variants.}
    \label{fig:arch}
    \vspace{-1.5em}
\end{figure*}

\noindent\textbf{IRControlNet.}
Given the reliable condition image $I_{RM}$, we then leverage the pre-trained Stable Diffusion for our generation module (GM). To conclude, it mainly involves three aspects: 1) condition encoding; 2) condition network; 3) feature modulation. Our IRControlNet has explored in-depth and provided effective modules for addressing each of them. The architecture is illustrated in Fig. \ref{fig:arch}.

\textbf{1) condition encoding.}
In IRControlNet, we utilize the pretrained and fixed VAE encoder $\mathcal{E}$ to encode the condition image $I_{RM}$ into the latent space for condition encoding:  $c_{RM}=\mathcal{E}(I_{RM})$,
where $c_{RM}$ is obtained condition latent. Since the VAE is trained on large-scale datasets, the obtained $c_{RM}$ is capable of preserving sufficient image information. 

\textbf{2) condition network.}
As for condition network, we follow ControlNet \cite{controlnet} and make a trainable copy of the pre-trained UNet encoder and middle block (denoted as $\mathbf{F}_{cond}$), 
which receives condition information and then outputs control signals. This copy strategy provides a good weight initialization for condition network.  
Then, we use the concatenation of the condition $c_{RM}$ and the noisy latent $z_t$ at time $t$ as input of $\mathbf{F}_{cond}$, which is denoted as $ z_t^\prime=cat(z_t, c_{RM})$. As the concatenation operator $cat(\cdot)$ will increase the channel number, we introduce a few parameters to the first layer of $\mathbf{F}_{cond}$ and initialize them to zero. This zero initialization functions similarly to zero convolution in ControlNet, which is to avoid random noise as gradients in the early stage of training. 

\textbf{3) feature modulation.} 
The previous condition network outputs multi-scale features, which will be used to modulate the intermediate features of the frozen UNet denoiser. Following ControlNet, we only modulate the middle block features and the skipped features through addition operation. Besides, zero convolutions are employed to connect the condition network with the fixed UNet denoiser for improving stability of model training.

During training, only the parameters of condition network and feature modulation will be updated. Specifically, we aim to minimize the following latent diffusion objective:
\begin{equation}
    \mathcal{L}_{GM} = \mathbb{E}_{z_t, c, t, \epsilon, c_{RM}}[||\epsilon-\epsilon_{\theta}(z_t, c, t, c_{RM})||^2_2],
\end{equation}
where the obtained result in this stage is denoted as $I_{GM}$.

\noindent\textbf{Discussion.} In this part, we aim to validate IRControlNet to be a solid backbone as a generation module in BIR tasks. Specifically, we construct four model variants (see Fig. \ref{fig:arch}) to obtain a comprehensive empirical analysis of the crucial components in IRControlNet.

\textbf{Variant 1.} Regarding condition encoding, we replace IRControlNet's condition encoder $\mathcal{E}$ by a tiny trained-from-scratch network, consisting of several stacked convolution layers and one zero convolution at the end. The encoded condition is added to the output features from the first layer of condition network. This model variant is identical to ControlNet.

\textbf{Variant 2.} Regarding condition network, we remove noisy $z_t$ and only use the condition latent $c_{RM}$ as the condition network input.

\textbf{Variant 3.} Regarding condition network, we do not copy the original weights from UNet denoiser but train the condition network from random initialization.

\textbf{Variant 4.} Regarding feature modulation, we control the middle block features and decoder features instead of skipped ones.

% The experimental results are presented in Section \ref{model_arch_result}. The best architecture we found is denoted as IRControlNet.
The comparison of Variant 1 (or ControlNet) and IRControlNet is in Fig.\ref{fig:ablation} (right) and Table \ref{tab:ablation_IRControlNet}. We observe that Variant 1 cannot maintain the original color of input LQ images, and the quantitative results are significantly worse than IRControlNet in PSNR (3dB$\downarrow$ on average). This observation reveals that condition encoding plays a vital role in controlling latent diffusion prior for IR tasks. 

\begin{figure}
    \centering
    \includegraphics[width=\linewidth]{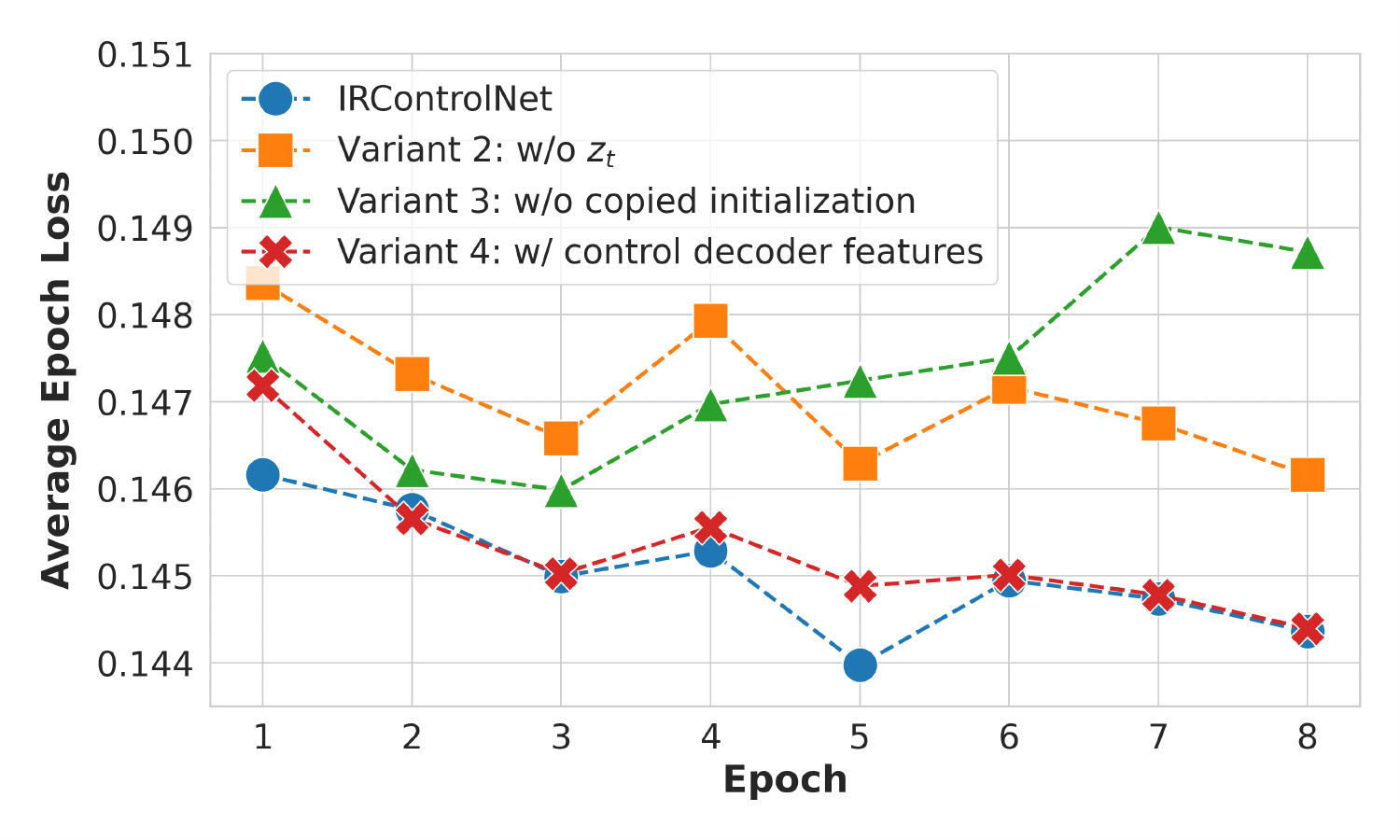}
    \vspace{-2em}
    \caption{The training loss curves of IRControlNet and Variant 2,3,4 on ImageNet1k dataset under the same training setting.}
    \label{fig:loss}
\end{figure}

\begin{table}
    \centering
    \resizebox{1.0\linewidth}{!}{
        % \begin{table}[]
% \begin{tabular}{c|ccc}
% \hline
% Variants                                & PSNR↑   & SSIM↑  & MANIQA↑ \\ \hline
% IRControlNet                            & 22.9865 & 0.5200 & 0.2689  \\
% \textbf{Variant 2:} w/o $z_t$                     & 23.1461 & 0.5398 & 0.2611  \\
% \textbf{Variant 3:} w/o copied initialization  & 22.8818 & 0.5192 & 0.2384  \\
% \textbf{Variant 4:} w/ control decoder features & 22.9721 & 0.5203 & 0.2686  \\ \hline
% \end{tabular}
% \end{table}

\begin{tabular}{c|ccc}
\hline
Variants                                & PSNR↑   & SSIM↑  & MANIQA↑ \\ \hline
IRControlNet                            & 22.9865 & 0.5200 & 0.2689  \\
\textbf{2}) w/o $z_t$                     & 23.1461 & 0.5398 & 0.2611  \\
\textbf{3}) w/o copied initialization  & 22.8818 & 0.5192 & 0.2384  \\
\textbf{4}) w/ control decoder features & 22.9721 & 0.5203 & 0.2686  \\ \hline
\end{tabular}}
    % \vspace{-0.6em}
    \captionof{table}{Quantitative comparisons of IRControlNet, Variant 2, 3 and 4 on ImageNet1k-Val with Real-ESRGAN\cite{realesrgan} degradation.}
    \label{tab:variants}
    \vspace{-1.5em}
\end{table}

The explanation might be that the image generation process is performed in the latent space, thus the condition should be projected to the same space. IRControlNet identifies it and cleverly uses the pretrained VAE encoder $\mathcal{E}$ for effective encoding and has achieved prominent improvement over ControlNet.

Next, we compare our IRControlNet with Variant 2,3,4 in both training and testing aspects.
% The training loss curves are in Figure \ref{fig:loss}, while the quantitative comparison is in Table \ref{tab:variants}.
In Fig. \ref{fig:loss}, we observe that IRControlNet achieves the fastest model convergence among all model variants, showing its superiority of architecture design. For Variant 2 (w/o $z_t$), its training losses are consistently higher than those of IRControlNet in all epochs. This indicates that $z_t$ could facilitate convergence, as it makes the condition network aware of randomness at each timestep, thus improving the accuracy of model predictions.
From the quantitative comparison in Table \ref{tab:variants}, Variant 2 achieves the best performance in metrics that measure fidelity, but its IQA score is worse than IRControlNet. 
From qualitative results in Appendix, we find that Variant 2 usually produces smooth results without sufficient texture details.
% Since the condition network of Variant 2 outputs the same control signals for different $z_t$ at each timestep given one condition image, it is natural for Variant 2 to produce smooth results.
To conclude, $z_t$ in condition network can boost convergence and helps generate high-quality results, so it is important in generation module and should be incorporated. As shown in Fig. \ref{fig:loss}, Variant 3 struggles in training loss convergence. Besides, it achieves the worst performance in all metrics. 
% This can be attributed to the fact that the parameters of the condition network are relatively large. With random initialization, it requires a long time to converge and the training process will be fairly unstable. 
Therefore, a good weight initialization for condition network is crucial in the generation module. As for Variant 4, it achieves comparable convergence speed and quantitative results to IRControlNet, thus applying control to skipped features or decoder features has similar effects. However, the channel numbers of decoder features are about twice the ones of corresponding skipped features, which will introduce more parameters and computation for feature modulation. Therefore, IRControlNet's feature modulation on skipped features is fairly enough.

% \xq{Table \ref{tab:variants} presents the quantitative results. Variant 3 achieves the worst performance across all metrics since it has not converged yet. Compared with IRControlNet, Variant 2 achieves higher fidelity (in terms of PSNR) and worse quality (in terms of MANIQA). the reason is that the condition network of variant 2 is optimized without $z_t$ as input, thus it tends to output compromised control signals for different $z_t$, leading to a limited generative capability. Variant 4 demonstrates comparable performance with IRControlNet, which is aligned with the training loss. However, adding control to decoder features will introduce more parameters, thus IRControlNet chooses to control on skipped features.}

In conclusion, IRControlNet proves to be a solid backbone for generative module in BIR tasks, as its main components are crucial for either model convergence or performance. We have compared more model variants in Appendix and our conclusion still stands.

\subsection{Restoration Guidance}
\label{sec:method3}
\begin{figure*}[htbp]
    % \vspace{-1.8em}
    \centering
    \includegraphics[width=\linewidth]{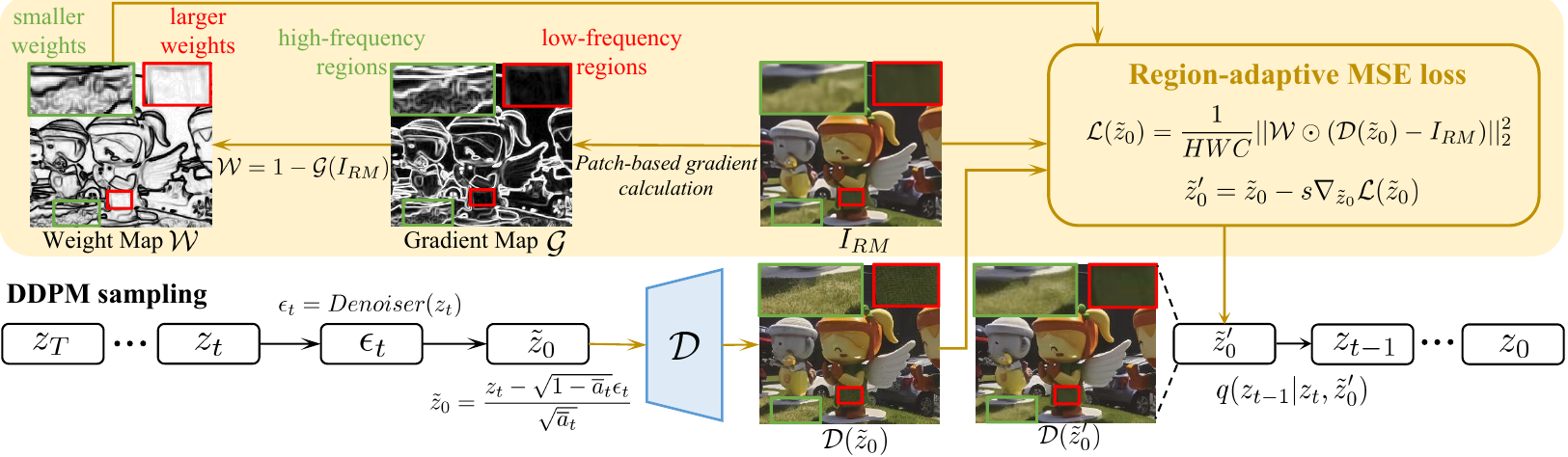}
    \vspace{-1.5em}
    \caption{Region-adaptive restoration guidance. 
    Given the high-fidelity guidance image $I_{RM}$, it aims to minimize the region-adaptive MSE loss between clean latent $\tilde{z}_0$ and $I_{RM}$ at each sampling step through gradient-descent algorithm.}
    \label{fig:guidance}
    \vspace{-1.5em}
\end{figure*}

% It is observed that the model generates unnecessary details in the windows and wall with, and MSE guidance tends to make the overall image blurry. Our restoration guidance considers the gradient signals of different regions, which allows for the removal of unnecessary details while maintaining the overall quality of the image.
Here we design a controllable module to achieve trade-off between \textit{quality} and \textit{fidelity}.
Note that users usually expect more generated details in high-frequency regions (\eg, textures, edges) but less generated content in flat regions (\eg, sky, wall).
% Although the above two-stage approach could already achieve high-quality results, a trade-off between \textit{realness} and \textit{fidelity} is still needed for various users’ preferences. Moreover, the generated results are sometimes with unpleasant artifacts due to the randomness of diffusion models (see the unusual texture details in the windows and wall in Figure \ref{fig:guidance_visualization}).
% \xq{Restoration guidance can be regarded as a guidance technique that is specially designed for BIR problem. Its main purpose is to achieve a trade-off between quality and fidelity.}
To this end, we present a region-adaptive restoration guidance, which guides the denoising process towards the restored result in stage \uppercase\expandafter{\romannumeral1} under a tunable guidance scale controlled by users. This restoration guidance is training-free and applied for every sampling step. The whole pipeline is in Fig. \ref{fig:guidance}.
% To address this demand/problem, we propose the restoration guidance to impose an explicit constraint on the generative process, obtaining more reliable results with higher fidelity.
% This approach is inspired by classifier guidance \cite{beatsgan}, which utilizes a classifier trained on noisy images to guide generation towards target class through. While in most cases, the pre-trained models that serve as classifiers are usually trained on clean images. To handle this situation, the works in \cite{blend, gdp} turn to modify the intermediate variable $\tilde{x}_0$ instead of $x_t$ to control the generation process of diffusion models. 
% Specifically, in the sampling process, we estimate a clean image $x_0$ from the noisy image $x_t$ by estimating the noise in $x_t$.
% \xq{The first guidance technique is classifier guidance \cite{beatsgan}. It uses class label as guidance, which is obviously insufficient for BIR tasks. Recent works\cite{diffuse_and_restore, gdp} utilizes LQ images as guidance, but they either require predefined degradation kernel or can only be applied to face images. On the contrary, restoration guidance can generalize well to general images under diversified degradation.}

At time $t$, the UNet denoiser first predicts the noise $\epsilon_t$ of the noisy latent $z_t$. Then the predicted noise $\epsilon_t$ is removed from $z_t$ to get the clean latent $\tilde{z}_0$:
\begin{align}
    \epsilon_t=\epsilon_{\theta}(z_t, c, t, c_{RM}),  
    \tilde{z}_0=\frac{z_t-\sqrt{1-\bar{\alpha}_t}\epsilon_t}{\sqrt{\bar{\alpha}_t}}.
\end{align}
In this stage, we aim to guide $\mathcal{D}(\tilde{z}_0)$ towards the high-fidelity condition $I_{RM}$. Thus, we propose a region-adaptive MSE loss function that applies between them in pixel space and update the clean latent $\tilde{z}_0$ with gradient descent algorithm. First, we compute the gradient magnitude by applying sobel operators. As pixels with strong gradient signals are very rare in an image, we then divide $I_{RM}$ into multiple non-overlapping patches and calculate patch-level gradient magnitude $\mathcal{G}(I_{RM})$ for better estimating the gradient density. More details can be found in Appendix.
Finally, we calculate a weight map by $\mathcal{W}=1-\mathcal{G}(I_{RM})$. 
And the MSE loss is adjusted by the weight map $\mathcal{W}$, and is defined as follows:
\begin{align}
    \mathcal{L}(\tilde{z}_0)=
    \frac{1}{HWC}||\mathcal{W}\odot(\mathcal{D}(\tilde{z}_0)-I_{RM})||^2_2, 
\end{align}
% Then, $z_{t-1}$ is sampled from the predefined distribution $q(z_{t-1}|z_t, \tilde{z}_0)$.
% Here we define a gradient-aware MSE loss between the predicted clean latent $\tilde{z}_0$ and condition latent $c_{reg}$ in image space:
% \begin{align}
%     \mathcal{L}(\tilde{z}_0, c_{reg})=
%     \frac{1}{HWC}||\mathcal{W}\odot(\mathcal{D}(\tilde{z}_0)-I_1)||^2_2,\\
%     \mathcal{W}=1-\mathcal{G}(I_{reg}),
% \end{align}
% where $N$ denotes the number of pixels in $I_{reg}$, $\mathcal{G}(I_{reg})$ is the normalized gradient magnitude of $I_{reg}$, and $\mathcal{W}$ is a weight map. $\mathcal{G}(I_{reg})$ represents the gradient intensity of each pixel in $I_{reg}$ and the detailed explanations are presented in supplementary file.
where $H, W, C$ denotes the spatial size of $I_{RM}$.
In this way, regions with weak gradients are assigned with larger weights, and vice versa. This indicates that low-frequency regions induce higher loss, thus they are influenced more by the high-fidelity condition $I_{RM}$. 
On the contrary, high-frequency regions are less affected and could maintain more generated content during sampling process. This analysis corresponds well to the illustration in Fig. \ref{fig:guidance}, in which the noisy content in flat regions is largely eliminated while the generated textures for glass are maintained well after restoration guidance. 

The gradient descent algorithm is applied for optimizing the region-adaptive MSE loss at each sampling step $t$ by the following equation:
\begin{equation}
    \tilde{z}^\prime_0=\tilde{z}_0 - s\nabla_{\tilde{z}_0}\mathcal{L}(\tilde{z}_0),
\end{equation}
where $s$ denotes the guidance scale, which can be used to control how much information is maintained from
the guidance image $I_{RM}$. For instance, larger guidance scale pushes $\mathcal{D}(\tilde{z}_0)$ closer to $I_{RM}$, indicating a higher fidelity. The whole algorithm of our restoration guidance is presented in Appendix.
\section{Experiments}
\subsection{Datasets, Implementation, Metrics}
\noindent\textbf{Datasets.}
We train DiffBIR on our filtered laion2b-en \cite{laion} dataset that contains around 15M high-quality images. All images are randomly cropped to $512\times512$ during training. 
% Specifically, images with a resolution higher than 1000, and then remove product images (e.g. shoes) that are far from the natural image distribution. 
% Finally, we obtain a subset which contains 20M \xq{TODO: verify the number of images} high-quality and high-diversity images.
We evaluate our method for 1) BSR task on three synthetic datasets: DIV2K-Val \cite{div2k}, DRealSR \cite{drealsr}, RealSR \cite{realsr},  and two real-world datasets: RealSRSet \cite{bsrgan} and our collected real47, 
% (a collection of low-quality images from the Internet that contains general images of diverse scenes, such as natural outdoor landscapes, old photos, architecture, humans from portraits to dense people crowds, plants, and animals, etc.). 
% we collect 47 images from the Internet, denoted as Real47. It contains general images of diverse scenes, such as natural outdoor landscapes, old photos, architecture, humans from portraits to dense people crowds, plants, and animals, etc.
2) BFR task on the real-world datasets LFW-Test \cite{gfpgan} and WIDER-Test \cite{codeformer}, and  
% The LFW-Test contains 1711 face images with mild degradation and the Wider-Test consists of 970 severely degradaded face images.
3) BID task on a mixed real-world dataset, which contains images from real3 \cite{scunet}, real9 \cite{scunet}, and RNI15 \cite{rni15}.
% For BID task, we evaluate our methods on a mixed real-world dataset consists of \cite{scunet}, real9 \cite{scunet} and RNI15 \cite{rni15}.

\noindent\textbf{Implementation.}
We train the restoration module for 150k iterations (batch size=96). 
Then, we adopt Stable Diffusion 2.1-base\footnote{\scriptsize Stable Diffusion v2.1: \url{https://github.com/Stability-AI/stablediffusion}} as the generative prior, and finetune the proposed IRControlNet for 80k iterations (batch size=256). Adam \cite{adam} is used as the optimizer. The learning rate is set to $10^{-4}$ for the first 30k iterations and then decreased to $10^{-5}$ for the following 50k iterations. The training process is conducted on $512\times512$ resolution with 8 NVIDIA A100 GPUs.

During inference, we replace our trained restoration module with off-the-shelf task-specific restoration models: BSRNet \cite{bsrgan}\footnote{\scriptsize{ \url{https://github.com/cszn/BSRGAN}}} 
for BSR, SwinIR \cite{swinir}\footnote{\scriptsize \url{https://github.com/zsyOAOA/DifFace}} 
used in DifFace \cite{difface} for BFR, and SCUNet-PSNR \cite{scunet}\footnote{\scriptsize \url{https://github.com/cszn/SCUNet}}
for BID. While the trained IRControlNet remains unchanged for all tasks.
The positive prompt is set to empty and we use texts like \textit{"low quality"}, \textit{"blurry"} as our negative prompt.
We set the restoration guidance scale to 0, 0.5, and 1 for comparisons on synthetic datasets. As for real-world scenarios, the restoration guidance scale is set to 0 for higher quality.
To accelerate the sampling process, we adopt a spaced DDPM sampling schedule \cite{iddpm} which requires 50 sampling steps.
For images larger than $512$, we directly feed them into DiffBIR. For images with sides $<512$, we first upsample them with the short side enlarged to 512, and then resize them back after restoration.

\noindent\textbf{Metrics}. For synthesized data, we adopt the traditional metrics: PSNR, SSIM, and LPIPS \cite{lpips}. To better evaluate the \textit{quality}, we include several no-reference image quality assessment (IQA) metrics: MANIQA\cite{maniqa}, MUSIQ \cite{musiq} and CLIP-IQA \cite{clipiqa}. For BFR, we employ the widely used perceptual metric FID \cite{fid}.

\subsection{Comparisons with State-of-the-Art Methods}
DiffBIR is compared with state-of-the-art 1) BSR methods: FeMaSR \cite{femasr}, DASR \cite{dasr}, Real-ESRGAN+ \cite{realesrgan}, BSRGAN \cite{bsrgan}, SwinIR-GAN \cite{swinir}, StableSR \cite{stablesr} and PASD \cite{pasd}, 
2) BFR methods: CodeFormer \cite{codeformer}, DifFace \cite{difface}, DMDNet \cite{dmdnet}, DR2 \cite{dr2}, GCFSR \cite{gcfsr}, GFP-GAN \cite{gfpgan}, GPEN \cite{gpen}, RestoreFormer++ \cite{restoreformer++},  VQFR \cite{vqfr} and PGDiff \cite{pgdiff}, 3) BID methods: CBDNet \cite{cbdnet}, DeamNet \cite{deamnet}, Restormer \cite{restormer}, SwinIR \cite{swinir} and SCUNet-GAN \cite{scunet}.

\begin{table*}[!h]
    \centering
    \resizebox{\linewidth}{!}{
% \begin{table}[]
\begin{tabular}{c|ccccc|cc|ccc}
\hline
          Metrics & \makecell{FeMaSR \cite{femasr}}  & \makecell{DASR \cite{dasr}}    & \makecell{Real-ESRGAN+ \cite{realesrgan}}                                          & \makecell{BSRGAN \cite{bsrgan}}                          & \makecell{SwinIR-GAN \cite{swinir}}                                            & \makecell{StableSR \cite{stablesr}} & \makecell{PASD \cite{pasd}}    & DiffBIR (s=0)                                          & DiffBIR (s=0.5)                                        & DiffBIR (s=1)                                          \\ \hline
PSNR↑     & 20.1303 & 21.2141 & 21.0348                                               & \cellcolor[HTML]{F2F2F2}21.4531 & 20.7488                                               & 21.2392  & 20.7838 & 20.5824                                                & \cellcolor[HTML]{F2F2F2}{\color[HTML]{0B5FD1} 21.5808} & \cellcolor[HTML]{F2F2F2}{\color[HTML]{FF0000} 21.9154} \\
SSIM↑     & 0.4451  & 0.4773  & \cellcolor[HTML]{F2F2F2}{\color[HTML]{0B5FD1} 0.4899} & 0.4814                          & \cellcolor[HTML]{F2F2F2}0.4844                        & 0.4790   & 0.4727  & 0.4277                                                 & 0.4794                                                 & \cellcolor[HTML]{F2F2F2}{\color[HTML]{FF0000} 0.4986}  \\
LPIPS↓    & 0.3971  & 0.4479  & \cellcolor[HTML]{F2F2F2}{\color[HTML]{0B5FD1} 0.3921} & 0.4095                          & \cellcolor[HTML]{F2F2F2}{\color[HTML]{FF0000} 0.3907} & 0.3993   & 0.4353  & 0.3939                                                 & \cellcolor[HTML]{F2F2F2}0.3935                         & 0.4263                                                 \\
MUSIQ↑    & 62.7855 & 58.1591 & 64.6389                                               & 62.9271                         & \cellcolor[HTML]{F2F2F2}65.4945                       & 57.8069  & 63.8094 & \cellcolor[HTML]{F2F2F2}{\color[HTML]{FF0000} 73.1019} & \cellcolor[HTML]{F2F2F2}{\color[HTML]{0B5FD1} 68.6657} & 61.1476                                                \\
MANIQA↑   & 0.1443  & 0.1531  & 0.2238                                                & 0.1833                          & 0.2061                                                & 0.1648   & 0.2354  & \cellcolor[HTML]{F2F2F2}{\color[HTML]{FF0000} 0.3836}  & \cellcolor[HTML]{F2F2F2}{\color[HTML]{0B5FD1} 0.3146}  & \cellcolor[HTML]{F2F2F2}0.2466                         \\
CLIP-IQA↑ & 0.5674  & 0.5571  & 0.5905                                                & 0.5195                          & 0.5779                                                & 0.5541   & 0.6125  & \cellcolor[HTML]{F2F2F2}{\color[HTML]{FF0000} 0.7656}  & \cellcolor[HTML]{F2F2F2}{\color[HTML]{0B5FD1} 0.7158}  & \cellcolor[HTML]{F2F2F2}0.6347                         \\ \hline
\end{tabular}
% \end{table}
    }
    \vspace{-0.8em}
    \caption{Quantitative comparisons on synthetic dataset (DIV2K-Val) for BSR task. {\color{red}\textbf{Red}} and {\color{blue}blue} indicate the best and second best. The top 3 results are marked as \colorbox{mygray}{gray}.}
    \label{tab:bsr_syn_div2k}
    \vspace{-1em}
\end{table*}

\begin{table*}[!h]
    % \vspace{-0.5em}
    \centering
    \resizebox{\linewidth}{!}{
% \usepackage{multirow}
% \begin{table}[]
\begin{tabular}{c|c|ccccc|cc|c}
\hline
                            Datasets & Metrics          & FeMaSR \cite{femasr}                                                & DASR \cite{dasr}   & Real-ESRGAN+ \cite{realesrgan} & BSRGAN \cite{bsrgan}                                                & SwinIR-GAN \cite{swinir} & StableSR \cite{stablesr} & PASD \cite{pasd}                                                   & DiffBIR (s=0)                                          \\ \hline
                            & MUSIQ↑    & 64.6735                                               & 59.2695 & 63.2675      & \cellcolor[HTML]{F2F2F2}{\color[HTML]{0B5FD1} 67.6705} & 64.2512    & 64.8372  & \cellcolor[HTML]{F2F2F2}67.4052                        & \cellcolor[HTML]{F2F2F2}{\color[HTML]{FF0000} 69.4208} \\
                            & MANIQA↑   & 0.2142                                                & 0.1595  & 0.1963       & \cellcolor[HTML]{F2F2F2}0.2240                         & 0.2054     & 0.2083   & \cellcolor[HTML]{F2F2F2}{\color[HTML]{0B5FD1} 0.2370}  & \cellcolor[HTML]{F2F2F2}{\color[HTML]{FF0000} 0.3211}  \\
\multirow{-3}{*}{RealSRSet \cite{bsrgan}} & CLIP-IQA↑ & \cellcolor[HTML]{F2F2F2}{\color[HTML]{0B5FD1} 0.6879} & 0.5236  & 0.5772       & 0.6456                                                 & 0.6008     & 0.6418   & \cellcolor[HTML]{F2F2F2}0.6761                         & \cellcolor[HTML]{F2F2F2}{\color[HTML]{FF0000} 0.7637}  \\ \hline
                            & MUSIQ↑    & 68.9384                                               & 62.2026 & 68.1098      & \cellcolor[HTML]{F2F2F2}69.4741                        & 68.8467    & 68.3422  & \cellcolor[HTML]{F2F2F2}{\color[HTML]{0B5FD1} 70.9712} & \cellcolor[HTML]{F2F2F2}{\color[HTML]{FF0000} 73.1397} \\
                            & MANIQA↑   & \cellcolor[HTML]{F2F2F2}0.2347                        & 0.1454  & 0.2055       & 0.2063                                                 & 0.2217     & 0.2264   & \cellcolor[HTML]{F2F2F2}{\color[HTML]{0B5FD1} 0.2607}  & \cellcolor[HTML]{F2F2F2}{\color[HTML]{FF0000} 0.3682}  \\
\multirow{-3}{*}{real47}    & CLIP-IQA↑ & \cellcolor[HTML]{F2F2F2}0.6911                        & 0.5445  & 0.6382       & 0.6111                                                 & 0.6246     & 0.6574   & \cellcolor[HTML]{F2F2F2}{\color[HTML]{0B5FD1} 0.6913}  & \cellcolor[HTML]{F2F2F2}{\color[HTML]{FF0000} 0.7781}  \\ \hline
\end{tabular}
% \end{table}
    }
    \vspace{-0.8em}
    \caption{Quantitative comparisons on real-world datasets for BSR task. {\color{red}\textbf{Red}} and {\color{blue}blue} indicate the best and second best performance. The top 3 results are marked as \colorbox{mygray}{gray}.}
    \label{tab:bsr_real}
    \vspace{-1em}
\end{table*}

\begin{table*}[!h]
    % \vspace{-2em}
    \centering
    \resizebox{1.0\linewidth}{!}{
% \usepackage{multirow}
% \begin{table}[]
\begin{tabular}{c|c|cccccccccc|c}
\hline
                        Datasets & Metrics                 & \makecell{CodeFormer\cite{codeformer}} & \makecell{DifFace\cite{difface}}                                                & \makecell{DMDNet\cite{dmdnet}}                          & \makecell{DR2 \cite{dr2}} 
                        % & \makecell{DR2(VQFR)\cite{dr2}} 
                        & \makecell{GCFSR\cite{gcfsr}}   & \makecell{GFP-GAN\cite{gfpgan}}                                                & \makecell{GPEN\cite{gpen}}                                                   & \makecell{RestoreFormer++\cite{restoreformer++}}                                       & \makecell{VQFR\cite{vqfr}}                            & \makecell{PGDiff\cite{pgdiff}}                                                 & DiffBIR (s=0)                                          \\ \hline
                        & MUSIQ↑           & 75.4830    & 70.4957                                                & 73.4027                         & 67.5357       & 71.3789 & \cellcolor[HTML]{F2F2F2}76.3779                       & \cellcolor[HTML]{F2F2F2}{\color[HTML]{FF0000} 76.6210} & 72.2492                                               & 74.3847                         & 72.2175                                                & \cellcolor[HTML]{F2F2F2}{\color[HTML]{0B5FD1} 76.4206} \\
                        & MANIQA↑          & 0.3188     & 0.2692                                                 & 0.2973                          & 0.2830          & 0.2790  & \cellcolor[HTML]{F2F2F2}{\color[HTML]{0B5FD1} 0.3688} & \cellcolor[HTML]{F2F2F2}0.3616                         & 0.3179                                                & 0.3280                          & 0.2927                                                 & \cellcolor[HTML]{F2F2F2}{\color[HTML]{FF0000} 0.4499}  \\
                        & CLIP-IQA↑        & 0.6890     & 0.5945                                                 & 0.6467                          & 0.5728         & 0.6143  & \cellcolor[HTML]{F2F2F2}{\color[HTML]{0B5FD1} 0.7196} & \cellcolor[HTML]{F2F2F2}0.7181                         & 0.7025                                                & 0.7099                          & 0.6133                                                 & \cellcolor[HTML]{F2F2F2}{\color[HTML]{FF0000} 0.7948}  \\
\multirow{-4}{*}{\makecell{LFW-Test \\ \cite{lfw}}}   & FID (ref. FFHQ)↓ & 52.8765    & 44.9201                                                & \cellcolor[HTML]{F2F2F2}43.5403 & 45.9420        & 52.6972 & 47.4717                                               & 51.9862                                                & 50.7309                                               & 50.1300                         & \cellcolor[HTML]{F2F2F2}{\color[HTML]{0B5FD1} 41.5814} & \cellcolor[HTML]{F2F2F2}{\color[HTML]{FF0000} 40.9065} \\ \hline
                        & MUSIQ↑           & 73.4081    & 65.2397                                                & 69.4709                         & 67.3163        & 69.9634 & \cellcolor[HTML]{F2F2F2}74.8308                       & \cellcolor[HTML]{F2F2F2}{\color[HTML]{FF0000} 75.6160} & 71.5155                                               & 71.4163                         & 66.0014                                                & \cellcolor[HTML]{F2F2F2}{\color[HTML]{0B5FD1} 75.3213} \\
                        & MANIQA↑          & 0.2971     & 0.2403                                                 & 0.2613                          & 0.2795          & 0.2803  & \cellcolor[HTML]{F2F2F2}{\color[HTML]{0B5FD1} 0.3508} & \cellcolor[HTML]{F2F2F2}0.3472                         & 0.2905                                                & 0.3060                          & 0.2406                                                 & \cellcolor[HTML]{F2F2F2}{\color[HTML]{FF0000} 0.4443}  \\
                        & CLIP-IQA↑        & 0.6984     & 0.5639                                                 & 0.6335                          & 0.5821        & 0.6266  & \cellcolor[HTML]{F2F2F2}0.7147                        & 0.7039                                                 & \cellcolor[HTML]{F2F2F2}{\color[HTML]{0B5FD1} 0.7171} & 0.7069                          & 0.5685                                                 & \cellcolor[HTML]{F2F2F2}{\color[HTML]{FF0000} 0.8085}  \\
\multirow{-4}{*}{\makecell{Wider-Test \\ \cite{codeformer}}} & FID (ref. FFHQ)↓ & 39.2517    & \cellcolor[HTML]{F2F2F2}{\color[HTML]{0B5FD1} 37.8440} & 38.9580                         & 40.1202       & 41.1986 & 41.3247                                               & 46.4419                                                & 45.4686                                               & \cellcolor[HTML]{F2F2F2}38.1675 & 40.2700                                                & \cellcolor[HTML]{F2F2F2}{\color[HTML]{FF0000} 35.8094} \\ \hline
\end{tabular}
% \end{table}
    }
    \vspace{-0.8em}
    \caption{Quantitative comparisons for BFR on real-world datasets. {\color{red}\textbf{Red}} and {\color{blue}blue} indicate the best and second best performance. The top 3 results are marked as \colorbox{mygray}{gray}.}
    \label{tab:bfr_real}
    \vspace{-1em}
\end{table*}

\begin{figure*}[!h]
    \centering
    \includegraphics[width=\linewidth]{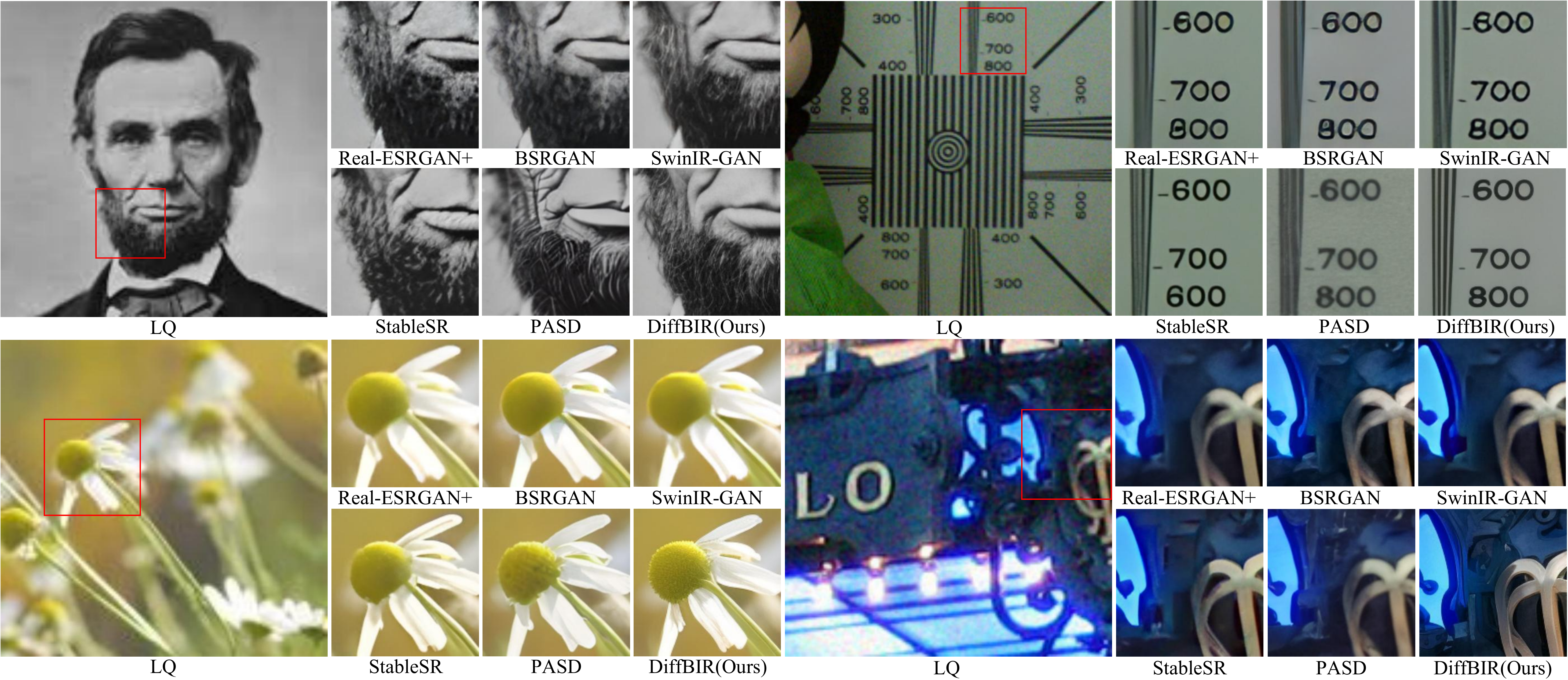}
    \vspace{-2em}
    \caption{Visual comparison of BSR methods on real-world datasets.}
    \label{fig:bsr_real}
    \vspace{-1em}
\end{figure*}

\begin{figure*}[!h]
    \centering
    \includegraphics[width=\linewidth]{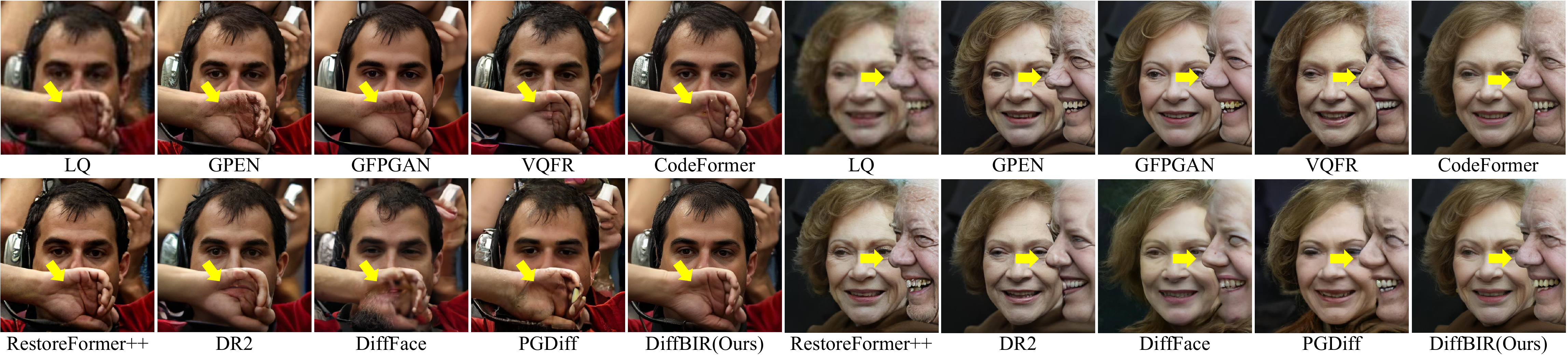}
    \vspace{-2em}
    \caption{Visual comparison of BFR methods on real-world datasets.}
    \label{fig:bfr_real}
\end{figure*}

\noindent{\textbf{BSR on synthetic datasets.}} Table \ref{tab:bsr_syn_div2k} presents quantitative comparisons on DIV2K-Val \cite{div2k} dataset. The LQ images are synthesized using the degradation model adopted in Real-ESRGAN \cite{realesrgan}.
It is observed that our DiffBIR ($s=0$) significantly outperforms all the baseline methods in terms of IQA metrics: MUSIQ, MANIQA, and CLIP-IQA. Moreover, our DiffBIR is able to obtain the best PSNR and SSIM when the restoration guidance scale is set to $1$, where the IQA metrics (MANIQA, CLIP-IQA) still rank top-3.
Users are recommended to control the restoration guidance scale to achieve a better balance between \textit{quality} and \textit{fidelity} (\eg, setting $s=0.5$).
% (Quantitative and visual comparisons on DRealSR and RealSR datasets are presented in supp.)
(Quantitative comparisons on DRealSR/RealSR and visual comparisons on DIV2K-Val are presented in Appendix.)

\noindent{\textbf{BSR on real-world datasets.}} We also provide the quantitative comparison on real-world datasets in Table \ref{tab:bsr_real}.
It is observed that our DiffBIR ($s=0$) obtains the best scores across all metrics on both the widely used RealSRSet \cite{bsrgan} and our collected Real47. 
This demonstrates DiffBIR's superiority in handling challenging real-world scenarios compared to the baseline methods.
As for visual comparison shown in Fig. \ref{fig:bsr_real}, DiffBIR is capable of producing sharper results than GAN-based methods, whose outputs tend to be over-smoothed. 
In contrast to diffusion-based methods, DiffBIR's restoration results are more realistic, such as the restored whiskers and lips, the pistil of flowers, texts, etc. 
More visual results can be found in Appendix.

\noindent{\textbf{BFR on real-world datasets.}}
We show the quantitative comparison on real-world datasets in Table \ref{tab:bfr_real}.
DiffBIR has achieved the highest FID score on both the LFW and Wider datasets, demonstrating its ability to generate more realistic faces. Regarding IQA metrics, DiffBIR also obtains the highest scores in CLIP-IQA and MANIQA, while the MUSIQ scores are close to the highest ones. 
Although the IRControlNet is not finetuned on face dataset (\eg, FFHQ), it outperforms all other baseline methods, indicating the excellent generalization ability of our proposed restoration pipeline.
The visual comparisons are shown in Fig. \ref{fig:bfr_real}. From the first example, it can be seen that only DiffBIR could restore the hand correctly, while other methods are influenced by facial priors thus distorting the hand area. In the second example, only DiffBIR successfully restores the side face, while other methods fail in restoring areas such as teeth, nose, and chin. Both two cases have demonstrated the superiority of using generative priors for general images rather than just face images.

\begin{figure}
    \centering
    \includegraphics[width=\linewidth]{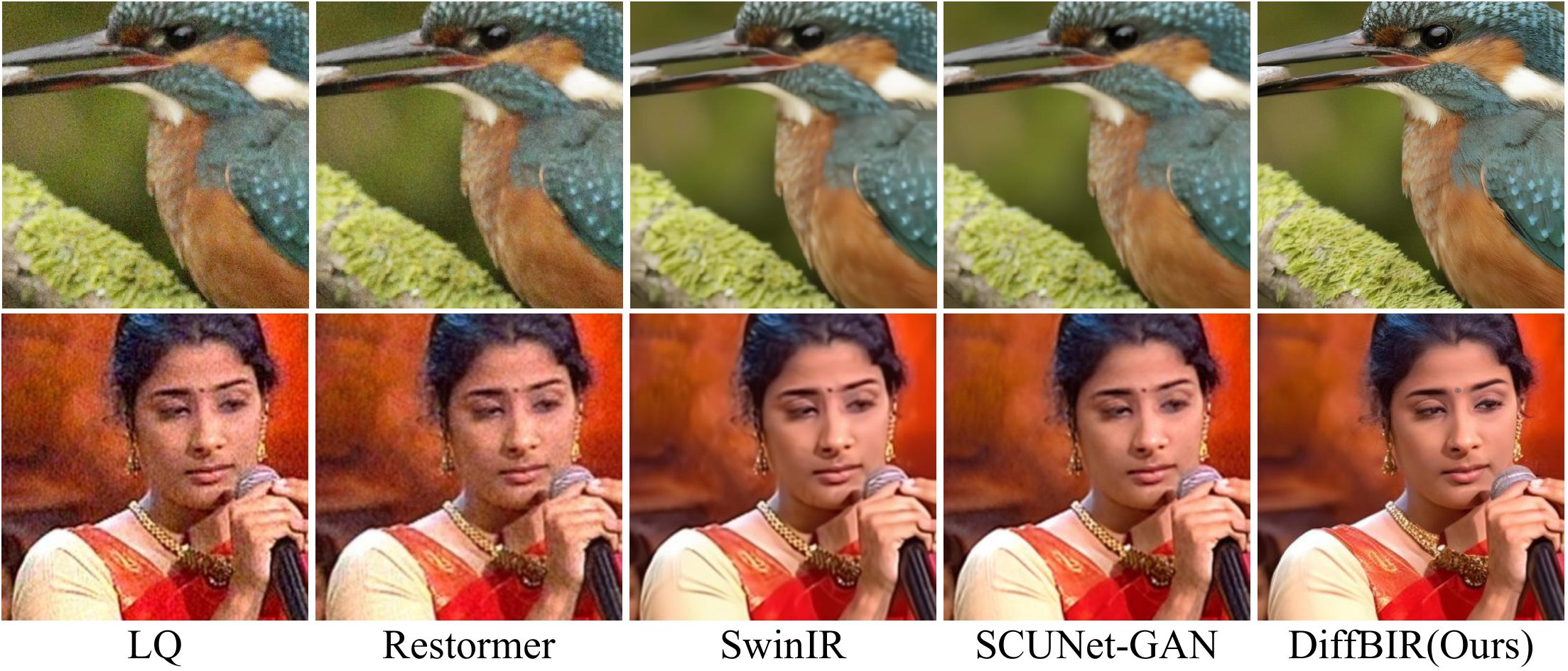}
    \vspace{-1.8em}
    \caption{Visual comparisons for BID on real-world datasets.}
    \label{fig:bid_real}
    \vspace{-0.7em}
\end{figure}

\noindent{\textbf{BID on real-world datasets.}} The quantitative comparisons are shown in Table \ref{tab:bid_real}. We can see that DiffBIR significantly outperforms the baseline methods across all metrics. This remarkable difference can be attributed to DiffBIR's introduction of powerful generative diffusion prior, which allows for effective high-quality image restoration.
Fig. \ref{fig:bid_real} illustrates visual comparisons between DiffBIR and baseline methods. It is observed that only DiffBIR can remove noise as well as generate realistic textures. Although SwinIR and SCUNet-GAN could successfully remove the noises, they produce smoothed results without vivid texture details.

\begin{table}
    \centering
    \resizebox{\linewidth}{!}{
% \begin{table}[]
\begin{tabular}{c|ccc}
\hline
        Methods & MUSIQ↑                                                 & MANIQA↑                                               & CLIP-IQA↑                                             \\ \hline
CBDNet \cite{cbdnet}        & 48.1149                                                & 0.1103                                                & \cellcolor[HTML]{F2F2F2}0.4709                        \\
DeamNet \cite{deamnet}       & 45.9942                                                & 0.0949                                                & 0.4391                                                \\
Restormer \cite{restormer}     & 47.4605                                                & 0.0927                                                & 0.3857                                                \\
SwinIR \cite{swinir}        & \cellcolor[HTML]{F2F2F2}55.0493                        & \cellcolor[HTML]{F2F2F2}0.1595                        & 0.4130                                                \\
SCUNet-GAN \cite{scunet}    & \cellcolor[HTML]{F2F2F2}{\color[HTML]{0B5FD1} 58.2170} & \cellcolor[HTML]{F2F2F2}{\color[HTML]{0B5FD1} 0.1822} & \cellcolor[HTML]{F2F2F2}{\color[HTML]{0B5FD1} 0.5045} \\
DiffBIR (s=0) & \cellcolor[HTML]{F2F2F2}{\color[HTML]{FF0000} 69.7278} & \cellcolor[HTML]{F2F2F2}{\color[HTML]{FF0000} 0.3404} & \cellcolor[HTML]{F2F2F2}{\color[HTML]{FF0000} 0.7420} \\ \hline
\end{tabular}
% \end{table}
    }
    \vspace{-0.5em}
    \captionof{table}{Quantitative comparisons on real-world datasets for BID task. {\color{red}\textbf{Red}} and {\color{blue}blue} indicate the best and second best. The top 3 results are marked as \colorbox{mygray}{gray}.}
    \label{tab:bid_real}
    \vspace{-0.7em}
\end{table}

\begin{figure}[htbp]
    \centering
    {\includegraphics[width=1\linewidth]{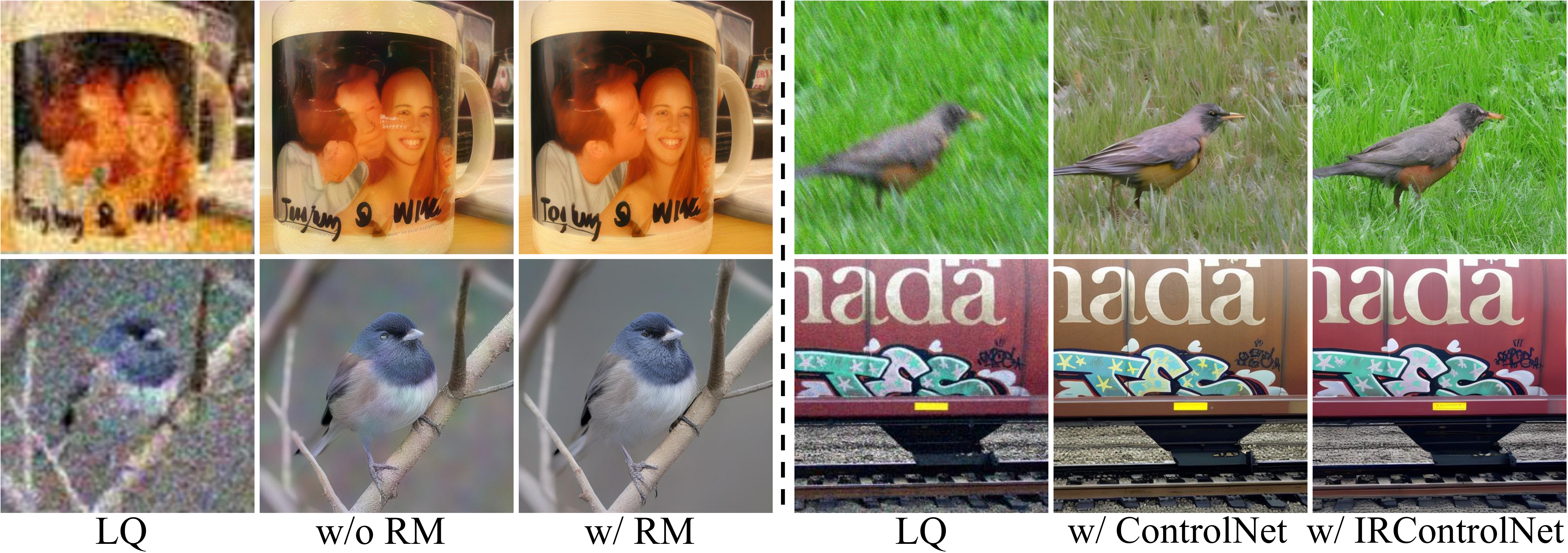}}
    \vspace{-1.8em}
    \caption{Visual comparison of ablation studies. (Left) DiffBIR w/o RM regards degradations as image content and performs poorly in fidelity maintaining; (Right) ControlNet\cite{controlnet} has a color shift problem which can be addressed by our IRControlNet.}
    \label{fig:ablation}
    \vspace{-1em}
\end{figure}

\subsection{Ablation Studies}

\label{sec:ablation1}
\noindent{\textbf{The Importance of Restoration Module.}} In this part, we investigate the significance of our proposed two-stage pipeline. Here, we remove the Restoration Module (RM) and directly finetune the diffusion model with synthesized training pairs.
From Table \ref{tab:ablation_RM}, the removal of the restoration module leads to a noticeable performance drop in all IQA and reference-based metrics on real-world and synthetic datasets. The visual comparison is presented in Fig.\ref{fig:ablation}(left).
From the first example, the one-stage model (w/o RM) causes severe distortion in facial generation. While the two-stage model could generate correct facial content. The second example shows that the one-stage model interprets the degradation as semantic information and produces a colorful background and unusual eye shapes. In contrast, the two-stage model produces more realistic results, demonstrating its superiority. 

\begin{table}[!h]
    % \vspace{-1.5em}
    \centering
    \resizebox{\linewidth}{!}{
% \usepackage{multirow}
% \begin{table}[]
\begin{tabular}{c|c|cc}
\hline
\multicolumn{1}{c|}{Datasets}      & 
\multicolumn{1}{c|}{Metrics}  & \multicolumn{1}{l}{w/o RM} & \multicolumn{1}{l}{w/ RM}     \\ \hline
                                  & MANIQA↑               & 0.2386                     & {\color[HTML]{FF0000} 0.2477}  \\
                                  & MUSIQ↑                & 62.5683                    & {\color[HTML]{FF0000} 64.7319} \\
\multirow{-3}{*}{RealSRSet \cite{bsrgan}}       & CLIP-IQA↑             & 0.6818                     & {\color[HTML]{FF0000} 0.7075}  \\ \hline
                                  & PSNR↑                 & 22.8481                    & {\color[HTML]{FF0000} 23.0078} \\
                                  & SSIM↑                 & 0.5039                     & {\color[HTML]{FF0000} 0.5198}  \\
\multirow{-3}{*}{ImageNet-Val-1k \cite{imagenet}} & LPIPS↓                & 0.4076                     & {\color[HTML]{FF0000} 0.4026}  \\ \hline
\end{tabular}
% \end{table}
    }
    \vspace{-0.5em}
    \caption{Ablation study on RM. The best results are denoted as \color{red}{Red}.}
    \label{tab:ablation_RM}
    \vspace{-1em}
\end{table}

\begin{table}[htbp]
    % \vspace{-2em}
    \centering
    \resizebox{1\linewidth}{!}{% Please add the following required packages to your document preamble:
% \begin{table}[]
\begin{tabular}{l|cccc}
\hline
                & Set14 \cite{set14}                          & BSD100 \cite{bsd100}                        & manga109 \cite{manga109}                      & ImageNet-Val-1k \cite{imagenet}                \\ \hline
w/ ControlNet   & 20.9435                        & 22.4923                        & 20.2692                        & 22.2874                        \\
w/ IRControlNet & {\color[HTML]{FF0000} 23.5193} & {\color[HTML]{FF0000} 23.8778} & {\color[HTML]{FF0000} 23.2439} & {\color[HTML]{FF0000} 24.2534} \\ \hline
\end{tabular}
% \end{table}
}
    \vspace{-0.5em}
    \caption{Comparison of ControlNet and ours in PSNR. {\color{red}{Red}} denotes the best results.}
    \label{tab:ablation_IRControlNet}
    \vspace{-0.5em}
\end{table}

\begin{table}
    \centering
    \resizebox{\linewidth}{!}{
% \begin{table}[]
\begin{tabular}{c|ccc}
\hline
               Degradation  & MANIQA↑                       & MUSIQ↑                         & CLIP-IQA↑                     \\ \hline
RealESRGAN \cite{realesrgan} & 0.2351                        & 64.1718                        & 0.6936                        \\
Ours & {\color[HTML]{FF0000} 0.2504} & {\color[HTML]{FF0000} 64.7319} & {\color[HTML]{FF0000} 0.7075} \\ \hline
\end{tabular}
% \end{table}
    }
    \vspace{-0.5em}
    \caption{Ablation study on degradation model evaluated on RealSRSet \cite{bsrgan}. {\color{red}{Red}} denotes the best results.}
    \label{tab:ablation_CodeFormer_deg}
    \vspace{-1em}
\end{table}

\noindent{\textbf{The Effectiveness of IRControlNet.}}
\label{sec:ablation2}
We compare our proposed IRControlNet with ControlNet \cite{controlnet} for BSR. As shown in Fig. \ref{fig:ablation}(right), ControlNet tends to output results with color shifts, as there is no explicit regularization of color consistency during training.  
The quantitative results presented in Table \ref{tab:ablation_IRControlNet} also show that our IRControlNet achieves higher PSNR scores than ControlNet.

\noindent{\textbf{The Effectiveness of Wide Degradation Range}}
\label{sec:ablation3}
In this work, we employ a classic degradation model with a wide degradation range to obtain conditions for training generation module, aiming to improve the overall generative capability. One commonly used degradation model for BSR is proposed by RealESRGAN \cite{realesrgan}. It adopts a very complex degradation process but uses much smaller degradation ranges.
Here we compare these two degradation models and present the quantitative comparison in Table \ref{tab:ablation_CodeFormer_deg}. It is observed that using our degradation model leads to better utilization of generative capabilities, thus enhancing the quality of the restored results.

% \vspace{-1em}
\section{Conclusion and Limitations}
We propose a unified framework for blind image restoration, named DiffBIR, which could achieve realistic restoration results by leveraging the prior knowledge of pre-trained Stable Diffusion. Extensive experiments have validated the superiority of DiffBIR over existing state-of-the-art methods for BSR, BFR, and BID tasks. Although our proposed DiffBIR has shown promising results, it requires 50 sampling steps to restore one low-quality image, which is computationally expensive. Besides, our two-stage restoration pipeline might be feasible for other BIR tasks, so more exploration can be conducted. 

{
    \small
    \bibliographystyle{ieeenat_fullname}
    \bibliography{main}
}

% \newpage
\section{Appendix}
% \subsection{Comparison of IRControlNet and More Variants}
\subsection{Comparison with More Variants}
\label{sec:supp_variants}
\begin{figure}[!h]
    % \vspace{-2em}
    \centering
    \includegraphics[width=\linewidth]{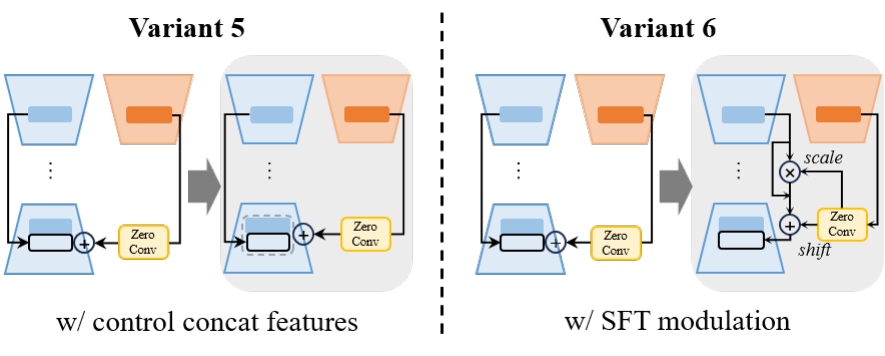}
    \vspace{-2em}
    \caption{Architectures of our IRControlNet and two model variants.}
    \label{fig:supp_more_variants}
    \vspace{-0.5em}
\end{figure}

\noindent\textbf{More Variants for IRControlNet.} For more comprehensive analysis, we construct another two variants. The architecture is illustrated in Fig. \ref{fig:supp_more_variants}.

\textbf{Variant 5.} Regarding feature modulation, we simultaneously control the middle block features, decoder features and skipped features. We use concat features for simplified denotation.

\textbf{Variant 6.} Regarding feature modulation, we use SFT layer\cite{sftgan} to modulate the intermediate features. Specifically as follows:
\begin{equation}
    SFT(\mathbf{F}|\bm{\gamma},\bm{\beta})=\mathbf{F}\odot(1+\bm{\gamma})+\bm{\beta}
\end{equation}
where $\mathbf{F}$ denotes feature maps, $\bm{\gamma}$ and $\bm{\beta}$ denotes the element-wise scale and shift transformation. Both $\bm{\gamma}$ and $\bm{\beta}$ are produced by zero-conv, thus they are initialized to zero at the beginning of training.

Table \ref{tab:supp_more_variants} presents the quantitative results. We can observe that both Variant 5 and 6 achieve better performance in terms of PSNR and SSIM while their MANIQA scores are worse than IRControlNet. Variant 5 applies more control on the pretrained model, which enhances the fidelity but damages generation quality. As for Variant 6, it utilizes SFT layer to modulate the skipped features. As SFT layer brings more precise control, which also improves the fidelity. In conclusion, both Variant 5,6 trade the quality for fidelity. IRControlNet achieves such a trade-off through restoration guidance and utilizes the add-on control to preserve most of the generation capability.

\begin{table}[htbp]
    % \vspace{-0.5em}
    \centering
    \resizebox{\linewidth}{!}{
        % \begin{table}[]
\begin{tabular}{c|ccc}
\hline
Variants                                & PSNR↑   & SSIM↑  & MANIQA↑ \\ \hline
IRControlNet                            & 22.9865 & 0.5200 & 0.2689  \\
Variant 5: w/ control concat   features & 23.0449 & 0.5261 & 0.2567  \\
Variant 6: w/ SFT modulation            & 22.9974 & 0.5292 & 0.2622  \\ \hline
\end{tabular}
% \end{table}
    }
    \vspace{-0.5em}
    \caption{Quantitative comparisons of IRControlNet, Variant 5 and 6 on ImageNet1k-Val with Real-ESRGAN\cite{realesrgan} degradation.}
    \label{tab:supp_more_variants}
    \vspace{-0.5em}
\end{table}

\begin{figure}[!h]
    % \vspace{-1.5em}
    \centering
    \includegraphics[width=\linewidth]{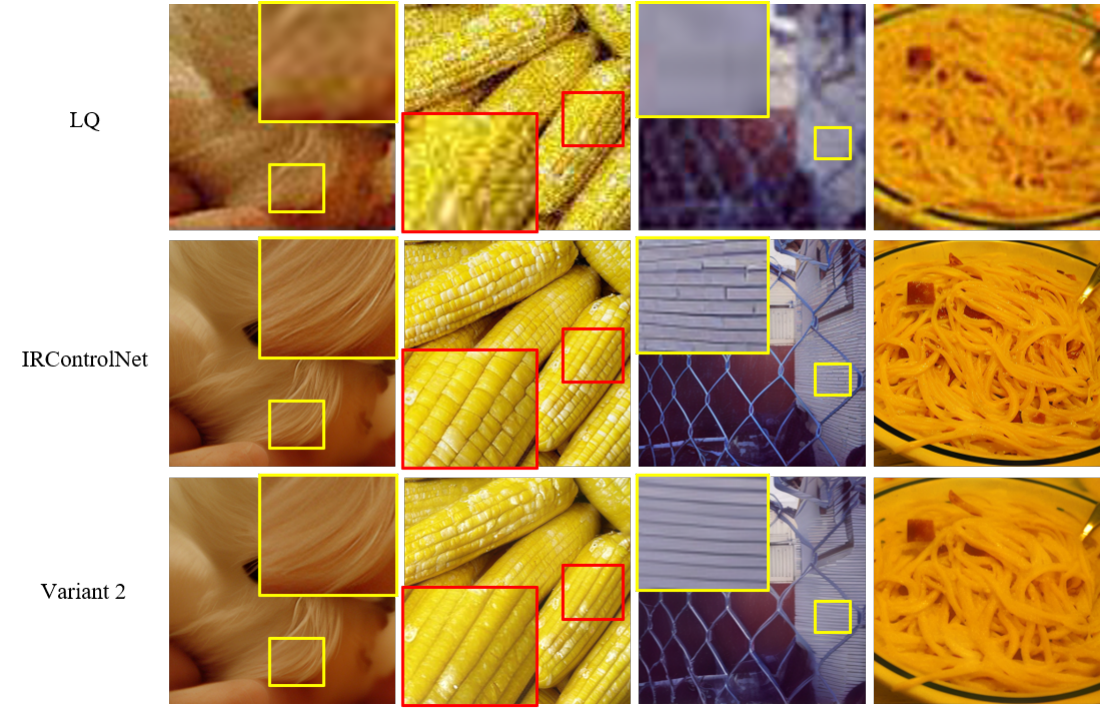}
    % \vspace{-1em}
    \caption{Visual comparisons of Variant 2 and IRControlNet.}
    \label{fig:supp_variant2}
    \vspace{-1em}
\end{figure}

\noindent\textbf{Qualitative Comparisons for Variant 2.} We present the visual comparisons for Variant 2 in Fig. \ref{fig:supp_variant2}. It can be observed that IRControlNet can generate more vivid textures while Variant 2 tends to produce over-smoothed results.

\subsection{More Details of Training RM}
\label{sec:supp_RM}
During the training of generation module, we follow \cite{difface} and modify a widely-used IR backbone, SwinIR \cite{swinir}, as our restoration module. Specifically, we utilize the pixel unshuffle \cite{pixelshuffle} operation to downsample the original low-quality input $I_{lq}$ with a scale factor of 8. For upsampling the deep features back to the original image space, we perform the nearest interpolation three times, and each interpolation is followed by one convolutional layer as well as one Leaky ReLU activation layer. This modified SwinIR will be trained on synthetic LQ-HQ image pairs.
Here we adopt a classic first-order degradation model to synthesize the LQ images.
\begin{align}
I_{lq} = \{[(I_{hq} \otimes k_{\sigma})_{\downarrow_r} + n_{\delta}]_{\texttt{JPEG}_q} \}_{\uparrow_r},
\end{align}
where the HQ image $I_{hq}$ is first convolved with a Gaussian kernel $k_{\sigma}$, followed by a downsampling of scale $r$. After that, additive Gaussian noise $n_{\delta}$ is added to the images, and then JPEG compression with quality factor $q$ is applied. Finally, the LQ image is resized back to the original size. Note that the downsampling and blurring contribute most to the information loss, thus we expand the degradation ranges of these two operations. Specifically, we randomly sample $\sigma$, $r$, $\delta$ and $q$ from \{0.1:12\}, \{1:12\}, \{0:15\}, \{30:100\}, respectively.
% \vspace{-1em}

\subsection{More Details about Restoration Guidance}
\label{sec:supp_guidance}
We provide a detailed explanation for our proposed restoration guidance in this section. Restoration guidance aims to achieve a trade-off between \textit{quality} and \textit{fidelity} through guiding the denoising process towards the high-fidelity $I_{RM}$ obtained in the first stage. At time $t$, the UNet denoiser first predicts the noise $\epsilon_t$ of the noisy latent $z_t$. Then the predicted noise $\epsilon_t$ is removed from $z_t$ to obatin the clean latent $\tilde{z}_0$ through the following equations:
\begin{align}
    \epsilon_t=\epsilon_{\theta}(z_t, c, t, c_{RM}), \\
    \tilde{z}_0=\frac{z_t-\sqrt{1-\bar{\alpha}_t}\epsilon_t}{\sqrt{\bar{\alpha}_t}}.
\end{align}
This indicates that we could modify the clean latent $\tilde{z}_0$ in each time step, and then sample $z_{t-1}$ according to the predefined distribution $q(z_{t-1}|z_t, \tilde{z}_0)$. 
In this way, we are able to achieve preferred restoration results without additional training.
To modify $\tilde{z}_0$, we define a region-adaptive MSE loss in image space:
\begin{align}
    \mathcal{L}(\tilde{z}_0)=
    \frac{1}{HWC}||\mathcal{W}\odot(\mathcal{D}(\tilde{z}_0)-I_{RM})||^2_2,\\
    \mathcal{W}=1-\mathcal{G}(I_{RM}),
\end{align}
where $H, W, C$ denotes the spatial size of $I_{RM}$, and $\mathcal{W}$ is a weight map. $\mathcal{G}(I_{RM})$ is the normalized gradient magnitude of $I_{RM}$, which represents the gradient intensity of each pixel in $I_{RM}$. To obtain $\mathcal{G}(I_{RM})$, we first calculate the gradient magnitude for each pixel in $I_{RM}$:
\begin{equation}
    M(I_{RM})=\sqrt{G_x(I_{RM})^2+G_y(I_{RM})^2}
\end{equation}
where $G_x$ and $G_y$ denotes the sobel operator in $x$ and $y$ axis, respectively.
As pixels with strong gradient signals are very rare in an image, we then use patch-level gradient signals for better estimate the gradient intensity. We divide $I_{RM}$ into multiple equal-sized non-overlapping patches as follows:
\begin{align}
    &\{I_{RM}^{(1)},I_{RM}^{(2)},...,I_{RM}^{(k)}, ...\}\\
    &\forall i, j,I_{RM}^{(i)}\cap{I_{RM}^{(j)}}=\emptyset,\bigcup_i I_{RM}^{(i)} = I_{RM}\nonumber
\end{align}
For patch $I_{RM}^{(k)}$, we calculate the sum of the gradient magnitudes of all pixels, and use the tanh function to map them into the range of $[0, 1)$:
\vspace{-0.5em}
\begin{equation}
    \vspace{-0.5em}
    S(I_{RM}^{(k)})=\tanh\left(\sum_{i,j} M_{i,j}(I_{RM})\right), (i,j)\in{I_{RM}^{(k)}}
\end{equation}
where $(i,j)$ denotes a pixel in patch $I_{RM}^{(k)}$. As $S(I_{RM}^{(k)})$ is closer to 1, the corresponding gradient signal is stronger, and vice versa. The final gradient magnitude can be formulated as below:
\begin{equation}
    \mathcal{G}_{i,j}(I_{RM})=\sum_{k}\mathbb{I}\left[(i,j)\in{I_{RM}^{(k)}}\right]S(I_{RM}^{(k)}),
\end{equation}
where $\mathbb{I}\left[(i,j)\in{P^{(k)}}\right]$ is an indicator function, denoting whether the pixel $(i,j)$ is located in the patch $I_{RM}^{(k)}$.
The whole algorithm is illustrated in Algorithm \ref{alg:supp_guidance}.

% \vspace{-1em}
\begin{algorithm}[htbp]
    \caption{Restoration guidance, given a diffusion model $\epsilon_{\theta}$, and the VAE's encoder $\mathcal{E}$ and decoder $\mathcal{D}$}
    \label{alg:supp_guidance}
    \begin{algorithmic}
    \Statex \textbf{Input:} Guidance image $I_{RM}$, text description $c$ (set to empty), diffusion steps $T$, gradient scale $s$
    \Statex \textbf{Output:} Output image $\mathcal{D}(z_0)$
    \Statex Sample ${z}_{T}$ from $\mathcal{N}(0, \mathbf{I})$
    \For{$t$ from $T$ to $1$}
        \State $\tilde{z}_0 \leftarrow \dfrac{z_t}{\sqrt{\bar{\alpha}_t}}-\dfrac{\sqrt{1-\bar{\alpha}_t}\epsilon_{\theta}(z_t, c, t, \mathcal{E}(I_{RM}))}{\sqrt{\bar{\alpha}_t}}$
        \State $\mathcal{W}=1-\mathcal{G}(I_{RM})$
        \State $\mathcal{L}(\tilde{z}_0)=\dfrac{1}{HWC}\left|\left|\mathcal{W}\odot(\mathcal{D}(\tilde{z}_0)-I_{RM})\right|\right|^2_2$
        \State Sample $z_{t-1}$ from $q(z_{t-1}|z_t, \tilde{z}_0-s\nabla_{\tilde{z}_0}\mathcal{L}(\tilde{z}_0))$
    \EndFor
    \State \textbf{return} $\mathcal{D}(z_0)$
    \end{algorithmic}
\end{algorithm}
\vspace{-1em}

\begin{table*}[htbp]
    % \vspace{-0.5em}
    \centering
    \resizebox{1.0\linewidth}{!}{
% \usepackage{multirow}
% \begin{table}[]
\begin{tabular}{c|c|ccccc|cc|ccc}
\hline
Datasets                  & Metrics   & FeMaSR \cite{femasr}  & DASR \cite{dasr}                                                   & Real-ESRGAN+ \cite{realesrgan} & BSRGAN \cite{bsrgan}                                                 & SwinIR-GAN \cite{swinir}                                            & StableSR \cite{stablesr} & PASD \cite{pasd}                                                  & DiffBIR (s=0)                                          & DiffBIR (s=0.5)                                        & DiffBIR (s=1)                   \\ \hline
                          & PSNR↑     & 23.1977 & \cellcolor[HTML]{F2F2F2}{\color[HTML]{FF0000} 26.3844} & 24.6878      & \cellcolor[HTML]{F2F2F2}{\color[HTML]{0B5FD1} 25.6903} & 25.3898                                               & 23.8669  & 24.8735                                               & 24.2037                                                & 24.9891                                                & \cellcolor[HTML]{F2F2F2}25.6238 \\
                          & SSIM↑     & 0.6239  & \cellcolor[HTML]{F2F2F2}{\color[HTML]{FF0000} 0.7271}  & 0.6705       & \cellcolor[HTML]{F2F2F2}0.6765                         & \cellcolor[HTML]{F2F2F2}{\color[HTML]{0B5FD1} 0.6962} & 0.6400   & 0.6529                                                & 0.5874                                                 & 0.6246                                                 & 0.6544                          \\
                          & LPIPS↓    & 0.2190  & \cellcolor[HTML]{F2F2F2}{\color[HTML]{FF0000} 0.1793}  & 0.2290       & 0.2308                                                 & \cellcolor[HTML]{F2F2F2}0.2057                        & 0.2355   & \cellcolor[HTML]{F2F2F2}{\color[HTML]{0B5FD1} 0.2016} & 0.2448                                                 & 0.2328                                                 & 0.2350                          \\
                          & MUSIQ↑    & 68.7458 & 66.0651                                                & 67.4608      & 68.9388                                                & 68.1393                                               & 69.2621  & \cellcolor[HTML]{F2F2F2}70.7670                       & \cellcolor[HTML]{F2F2F2}{\color[HTML]{FF0000} 72.3514} & \cellcolor[HTML]{F2F2F2}{\color[HTML]{0B5FD1} 71.5339} & 69.8821                         \\
                          & MANIQA↑   & 0.3073  & 0.2048                                                 & 0.2315       & 0.2309                                                 & 0.2375                                                & 0.2565   & 0.2889                                                & \cellcolor[HTML]{F2F2F2}{\color[HTML]{FF0000} 0.3915}  & \cellcolor[HTML]{F2F2F2}{\color[HTML]{0B5FD1} 0.3847}  & \cellcolor[HTML]{F2F2F2}0.3530  \\
\multirow{-6}{*}{DRealSR \cite{drealsr}} & CLIP-IQA↑ & 0.6327  & 0.5086                                                 & 0.5022       & 0.5328                                                 & 0.5244                                                & 0.5988   & 0.6151                                                & \cellcolor[HTML]{F2F2F2}{\color[HTML]{FF0000} 0.6878}  & \cellcolor[HTML]{F2F2F2}{\color[HTML]{0B5FD1} 0.6761}  & \cellcolor[HTML]{F2F2F2}0.6440  \\ \hline
                          & PSNR↑     & 23.1627 & \cellcolor[HTML]{F2F2F2}{\color[HTML]{FF0000} 25.5503} & 24.2400      & \cellcolor[HTML]{F2F2F2}{\color[HTML]{0B5FD1} 24.9717} & 24.6244                                               & 23.5627  & 24.5385                                               & 23.5237                                                & 24.2216                                                & \cellcolor[HTML]{F2F2F2}24.7531 \\
                          & SSIM↑     & 0.6534  & \cellcolor[HTML]{F2F2F2}{\color[HTML]{FF0000} 0.7183}  & 0.6793       & \cellcolor[HTML]{F2F2F2}0.6839                         & \cellcolor[HTML]{F2F2F2}{\color[HTML]{0B5FD1} 0.7051} & 0.6549   & 0.6694                                                & 0.5989                                                 & 0.6346                                                 & 0.6615                          \\
                          & LPIPS↓    & 0.2520  & \cellcolor[HTML]{F2F2F2}0.2397                         & 0.2556       & 0.2545                                                 & \cellcolor[HTML]{F2F2F2}{\color[HTML]{0B5FD1} 0.2340} & 0.2429   & \cellcolor[HTML]{F2F2F2}{\color[HTML]{FF0000} 0.2317} & 0.2646                                                 & 0.2544                                                 & 0.2565                          \\
                          & MUSIQ↑    & 66.1208 & 59.5565                                                & 66.7333      & 68.0673                                                & 67.0964                                               & 68.4594  & \cellcolor[HTML]{F2F2F2}70.0043                       & \cellcolor[HTML]{F2F2F2}{\color[HTML]{FF0000} 72.3909} & \cellcolor[HTML]{F2F2F2}{\color[HTML]{0B5FD1} 71.3969} & 69.5167                         \\
                          & MANIQA↑   & 0.2652  & 0.1713                                                 & 0.2243       & 0.2329                                                 & 0.2281                                                & 0.2407   & 0.2746                                                & \cellcolor[HTML]{F2F2F2}{\color[HTML]{FF0000} 0.3820}  & \cellcolor[HTML]{F2F2F2}{\color[HTML]{0B5FD1} 0.3792}  & \cellcolor[HTML]{F2F2F2}0.3504  \\
\multirow{-6}{*}{RealSR \cite{realsr}}  & CLIP-IQA↑ & 0.5925  & 0.4300                                                 & 0.4787       & 0.5233                                                 & 0.4920                                                & 0.5852   & 0.5822                                                & \cellcolor[HTML]{F2F2F2}{\color[HTML]{FF0000} 0.6868}  & \cellcolor[HTML]{F2F2F2}{\color[HTML]{0B5FD1} 0.6817}  & \cellcolor[HTML]{F2F2F2}0.6478  \\ \hline
\end{tabular}
% \end{table}
    }
    \vspace{-0.5em}
    \caption{Quantitative comparisons on synthetic datasets (DRealSR \cite{drealsr} and RealSR \cite{realsr}) for BSR task. {\color{red}\textbf{Red}} and {\color{blue}blue} indicate the best and second best performance. The top 3 results are marked as \colorbox{mygray}{gray}.}
    \label{tab:bsr_syn_drealsr_realsr}
    % \vspace{-4em}
\end{table*}

\begin{table*}[htbp]
    \centering
    \resizebox{1.0\linewidth}{!}{
        % \begin{table}[]
\begin{tabular}{c|cccccccc}
\hline
Metrics              & Real-ESRGAN+ \cite{realesrgan} & BSRGAN \cite{bsrgan} & SwinIR-GAN \cite{swinir} & FeMaSR \cite{femasr} & DASR \cite{dasr} & StableSR \cite{stablesr} & PASD \cite{pasd} & DiffBIR  \\ \hline
Inference Time (ms) & 46.19      & 46.42  & 126.44     & 89.01  & 12.69 & 19278.46 & 16951.08 & 10906.51 \\
Model Size (M)      & 16.69      & 16.69  & 11.71      & 34.05  & 8.06  & 1409.11  & 1675.76  & 1716.7   \\ \hline
\end{tabular}
% \end{table}
    }
    \vspace{-0.5em}
    \caption{Quantitative comparisons of inference efficiency and model complexity.}
    \label{tab:supp_infer_time}
    % \vspace{-2em}
\end{table*}

\subsection{More Quantitative and Qualitative Comparisons for BSR on Synthetic Datasets}
\label{sec:supp_bsr_syn}

\begin{figure*}[htbp]
    \centering
    \includegraphics[width=\linewidth]{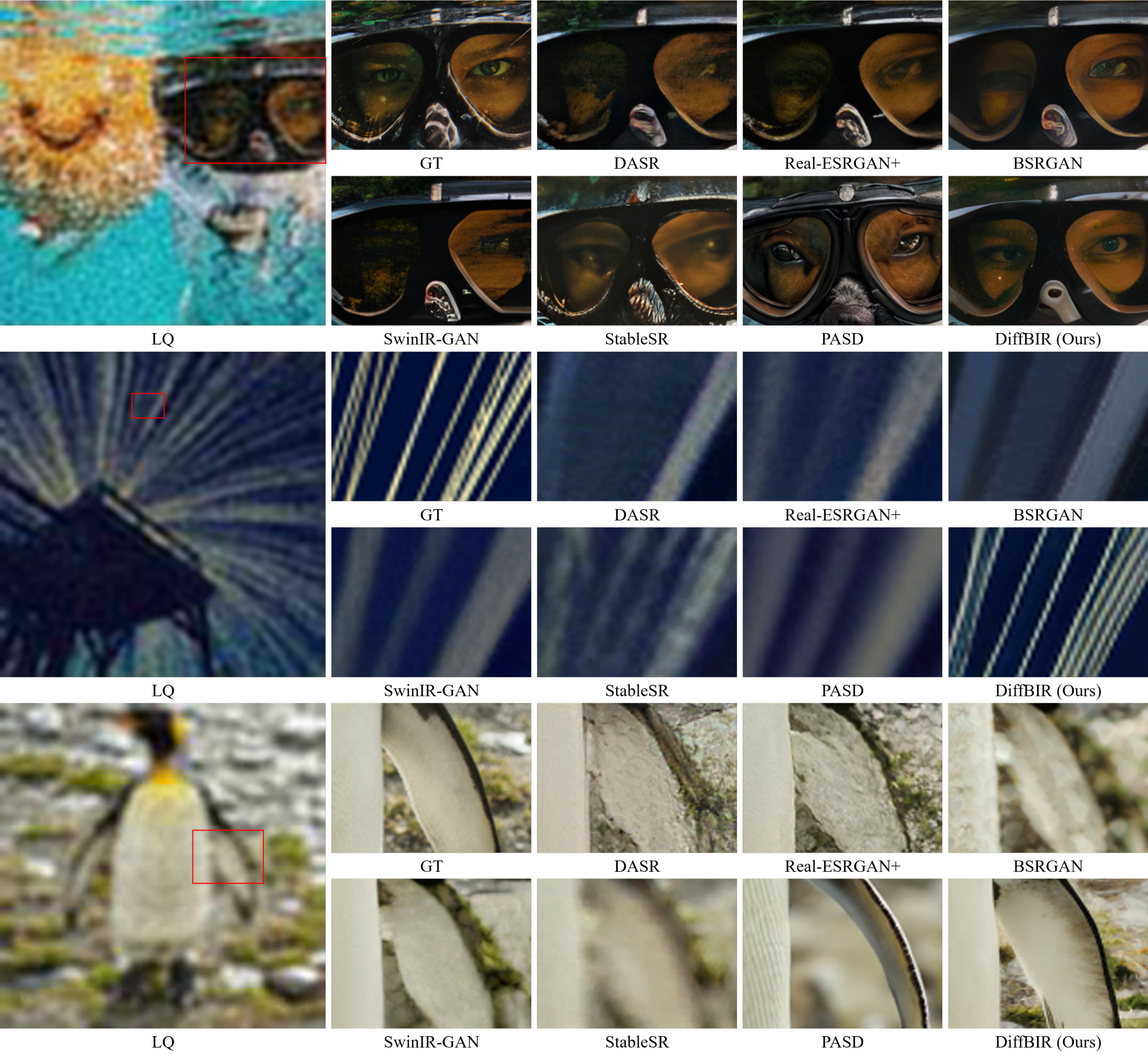}
    % \vspace{-2em}
    \caption{Visual comparisons of BSR methods on synthetic dataset (DIV2K-Val \cite{div2k}).}
    \label{fig:supp_bsr_syn}
\end{figure*}

The quantitative results on DRealSR \cite{drealsr} and RealSR \cite{realsr} are presented in Table \ref{tab:bsr_syn_drealsr_realsr}. The comparisons on these two datasets lead to similar observations. When the guidance scale $s$ is set to 0, DiffBIR significantly outperforms baseline methods in terms of all IQA metrics. When the guidance scale $s$ is set to 1, DiffBIR still surpasses the baseline methods in MANIQA and CLIP-IQA. As for evaluation in PSNR, DiffBIR performs better than diffusion-based methods and shows comparable performance to GAN-based methods, indicating that DiffBIR can achieve a good balance between \textit{quality} and \textit{fidelity}. Visual comparisons on DIV2K-Val\cite{div2k} are presented in Figure \ref{fig:supp_bsr_syn}. We can observe that only DiffBIR is able to produce restored results with correct semantic information. For example, it correctly recovers details such as the eyes behind the helmet, the lines of fireworks, and the wings of the penguin. GAN-based methods shows a lack of generation capability, thus producing over-smoothed results. In comparison, diffusion-based baseline methods are usually affected by the severe degradation and fail to generate correct semantics.

% \subsection{Quantitative Comparisons for Inference Efficiency}
\subsection{Quantitative Comparisons for Efficiency}
\label{sec:supp_inference_time}

We present a quantitative comparison regarding inference speed and model complexity for both diffusion-based and GAN-based methods in Table \ref{tab:supp_infer_time}. This comparison is performed on a super-resolution task with an input size of $128\times{128}$ and a scale factor of 4.
We conduct multiple inferences and calculate the average inference time.
% It can be observed that GAN-based methods are significantly faster than diffusion-based methods. While among diffusion-based methods, DiffBIR is the most efficient due to the fewer sampling steps and simpler architecture. In practical applications, we can employ some engineering methods to speed up inference, such as using half or lower precision.
It can be observed that DiffBIR is the most efficient among DM-based baselines. It's about 1.8x faster than StableSR and about 1.6x faster than PASD.
Although GAN-based methods are more efficient, they perform significantly worse than DM-based methods.
The development of diffusion models is extremely fast. There're works \cite{lcm, sdturbo} that can already achieve satisfactory generation performance with only 1$\sim$4 steps, thus the time-consuming problem can be solved in the future.

% \subsection{More Visual Comparisons on Real-world Datasets}
\subsection{More Real-world Visual Comparisons}
\label{sec:supp_real_world}
\begin{figure*}[!h]
    % \vspace{-1em}
    \centering
    \includegraphics[width=0.9\linewidth]{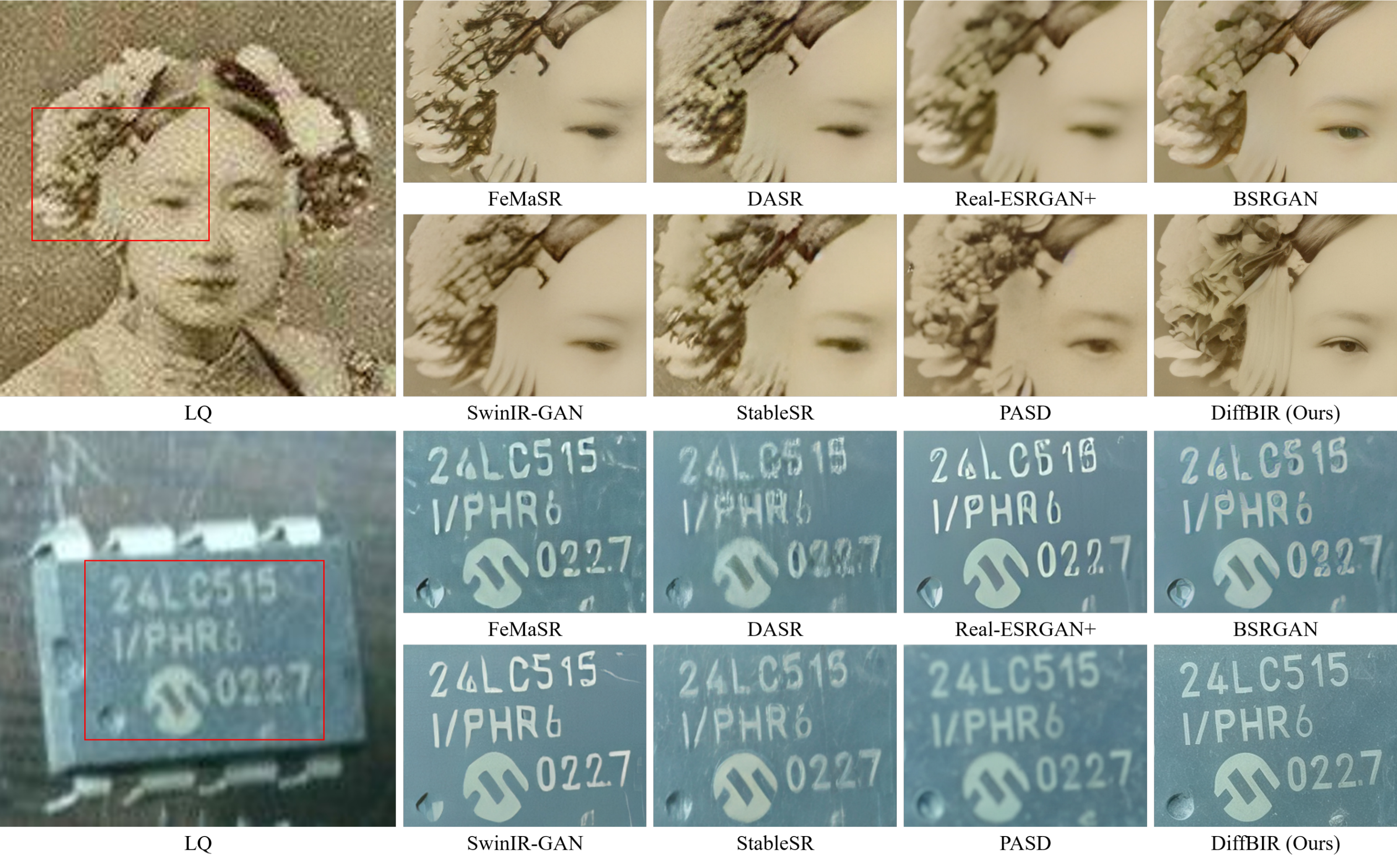}
    \vspace{-1em}
    \caption{More visual comparisons for BSR on real-world datasets.}
    \label{fig:supp_bsr_real}
    \vspace{-0.5em}
\end{figure*}

\begin{figure*}[!h]
    % \vspace{-4em}
    \centering
    \includegraphics[width=0.9\linewidth]{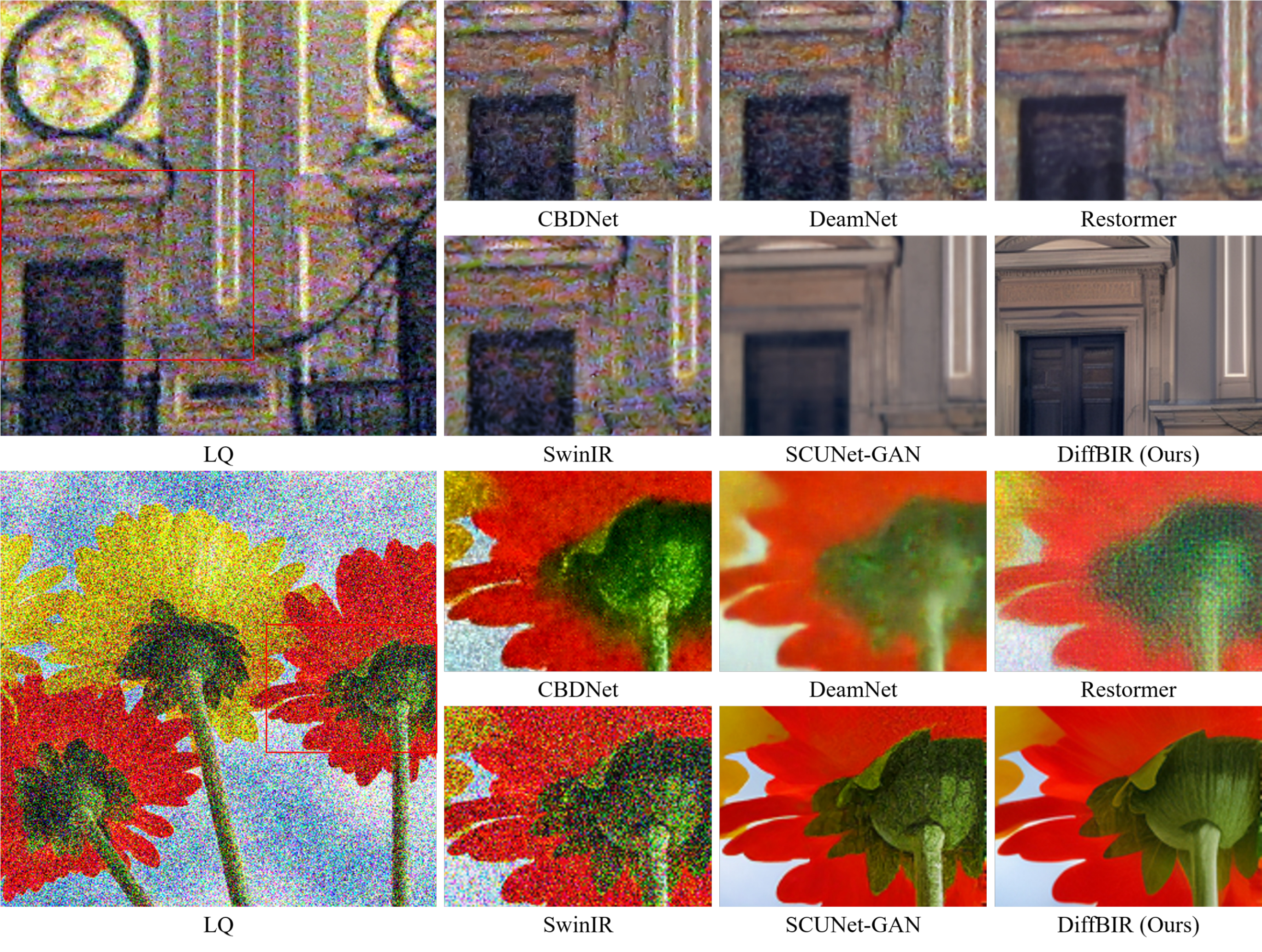}
    \vspace{-1em}
    \caption{More visual comparisons for BID on real-world datasets.}
    \label{fig:supp_bid_real}
\end{figure*}

\begin{figure*}[!h]
    \centering
    \includegraphics[width=\linewidth]{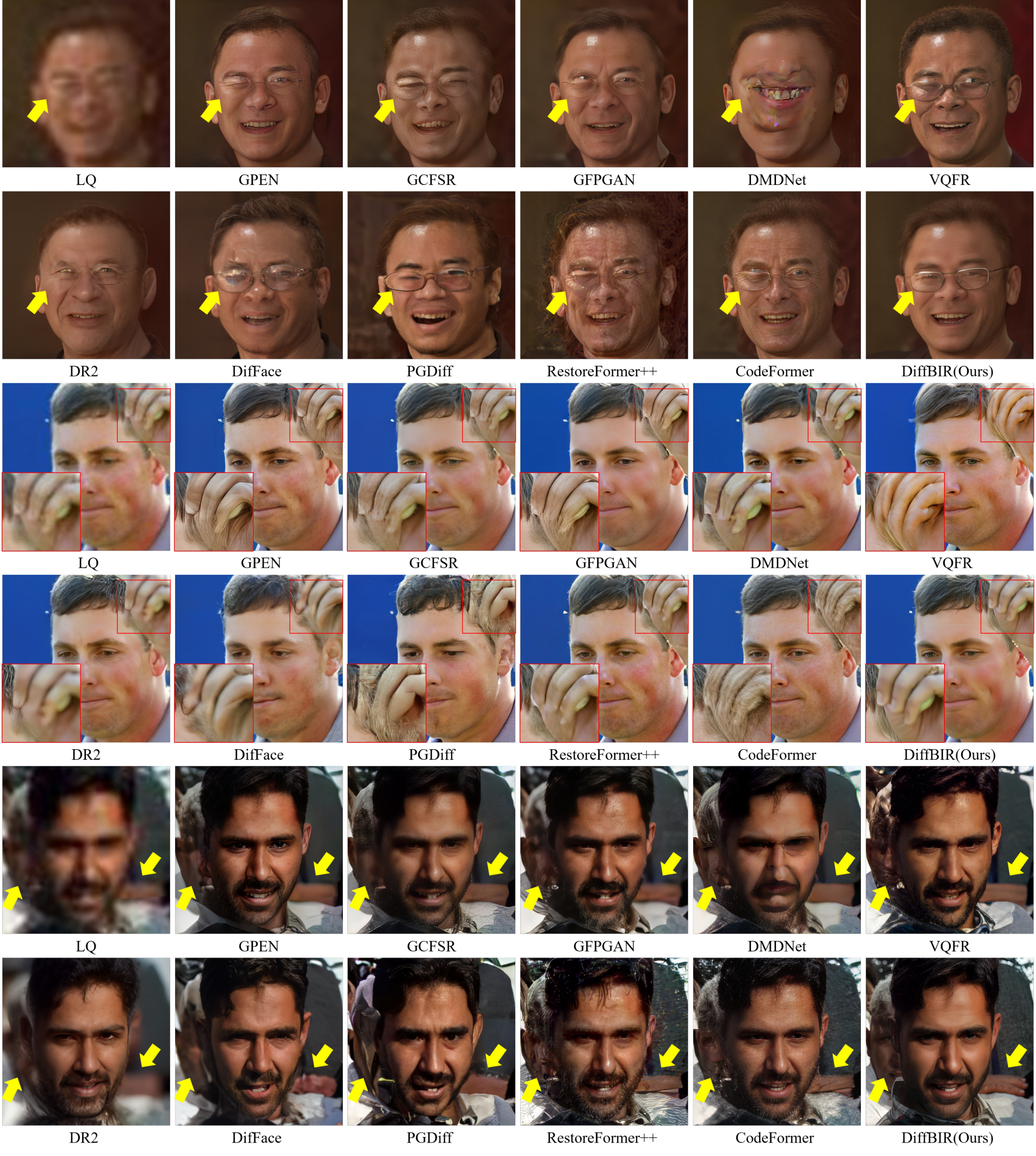}
    \vspace{-1em}
    \caption{\small More visual comparisons for BFR on real-world datasets.}
    \label{fig:supp_bfr_real}
\end{figure*}

We provide more visual comparisons for BSR, BID, BFR tasks in Figure \ref{fig:supp_bsr_real}, Figure \ref{fig:supp_bid_real} and Figure \ref{fig:supp_bfr_real}, respectively.

\end{document}